\documentclass[10pt]{article} % For LaTeX2e
\usepackage[preprint]{tmlr}
\usepackage{graphicx}
\usepackage{svg}
\usepackage{amsmath}
\usepackage{amsthm}
\usepackage{booktabs}
\usepackage{algorithm}
\usepackage{algpseudocode}
\usepackage{multirow}
\usepackage{multicol}
\usepackage{amsfonts}
\usepackage{amssymb}
\usepackage{subcaption}

\usepackage{amsmath,amsfonts,bm}

\def\eqref#1{equation~\ref{#1}}
\def\1{\bm{1}}

\DeclareMathAlphabet{\mathsfit}{\encodingdefault}{\sfdefault}{m}{sl}
\SetMathAlphabet{\mathsfit}{bold}{\encodingdefault}{\sfdefault}{bx}{n}

\newtheorem{theorem}{Theorem}

\newtheorem{definition}[theorem]{Definition}
\newtheorem{proposition}[theorem]{Proposition}
\newtheorem{corollary}[theorem]{Corollary}

\usepackage{hyperref}
\usepackage{url}

\title{Riemannian Neural Geodesic Interpolant}

\author{\name Jiawen Wu \email wujiawen19@mails.ucas.ac.cn \\
      \addr Academy of Mathematics and Systems Science, Chinese Academy of Sciences\\
      University of Chinese Academy of Sciences
      \AND
      \name Bingguang Chen \email bgchen@fjnu.edu.cn \\
      \addr School of Mathematics and Statistics, Fujian Normal University
      \AND
      \name Yuyi Zhou \email zhouyuyi23@mails.ucas.ac.cn\\
      \addr Academy of Mathematics and Systems Science, Chinese Academy of Sciences\\
      University of Chinese Academy of Sciences
      \AND
      \name Qi Meng \email meq@amss.ac.cn\\
      \addr Academy of Mathematics and Systems Science, Chinese Academy of Sciences
      \AND 
      \name Rongchan Zhu \email zhurongchan@126.com\\
      \addr Beijing Institute of Technology
      \AND 
      \name Zhi-Ming Ma \email mazm@amt.ac.cn\\
      \addr Academy of Mathematics and Systems Science, Chinese Academy of Sciences
      }

\def\month{MM}  % Insert correct month for camera-ready version
\def\year{YYYY} % Insert correct year for camera-ready version
\def\openreview{\url{https://openreview.net/forum?id=XXXX}} % Insert correct link to OpenReview for camera-ready version

\begin{document}

\maketitle

\begin{abstract}
Stochastic interpolants are efficient generative models that bridge two arbitrary probability density functions in finite time, enabling flexible generation from the source to the target distribution or vice versa. These models are primarily developed in Euclidean space, and are therefore limited in their application to many distribution learning problems defined on Riemannian manifolds in real-world scenarios. In this work, we introduce the \emph{Riemannian Neural Geodesic Interpolant} (RNGI) model, which interpolates between two probability densities on a Riemannian manifold along the stochastic geodesics, and then samples from one endpoint as the final state using the continuous flow originating from the other endpoint. We prove that the temporal marginal density of RNGI solves a transport equation on the Riemannian manifold. After training the model's the neural velocity and score fields, we propose the \emph{Embedding Stochastic Differential Equation} (E-SDE) algorithm for stochastic sampling of RNGI. E-SDE significantly improves the sampling quality by reducing the accumulated error caused by the excessive intrinsic discretization of Riemannian Brownian motion in the classical \emph{Geodesic Random Walk} (GRW) algorithm. We also provide theoretical bounds on the generative bias measured in terms of KL-divergence. Finally, we demonstrate the effectiveness of the proposed RNGI and E-SDE through experiments conducted on both collected and synthetic distributions on $\mathbb{S}^2$ and $\mathbf{SO}(3)$.
\end{abstract}

\section{Introduction}

Stochastic interpolants \citep{SIs,SI,SI2} are a type of generative models that use continuous-time stochastic processes with smooth path to precisely bridge two arbitrary probability density functions in finite time. These interpolants theoretically follow a transport equation from source distribution to target distribution, and their structures leads to simpler optimization loss function compared to the normalizing flow approach in the stage of training and leverages efficient numerical schemes of stochastic differential equations in the stage of sampling. It is regarded as a unifying framework for flow-based and diffusion-based generative models and demonstrates its efficiency on practical density estimation or image generation tasks.

Stochastic interpolants are primarily developed in the Euclidean space, but in many real applications, it's more mathematically rigorous to regard the livespace of data as a curved Riemannian manifold instead of Euclidean space. For example, the data distributed on earth lives in the spherical surface (i.e., $\mathbb{S}^2$) in geoscience \citep{TKDE}, and the molecular data satisfying 3d rotational equivariance live in $\mathbf{SO}(3)$ group in molecular biology\citep{Nature}. Generative modeling for these data with geometric structure benefits from the specialized geometric model, which is established in the Riemannian rather than the Euclidean sense to approximate stochastic law latently dominated by the curvature of space more accurately. Without the prior-encoding of specific Riemannian structures, conventional Euclidean models fail to adequately capture the intrinsic spatial information embedded in data when directly applied to non-Euclidean tasks, e.g., assigning a non-zero probability to regions outside the desired manifold. 

%As the prosperous development of Large Language Models as GPT, Dall-E and Sora,
%generative modeling has become one of the most central and attractive topics in machine learning community. Diffusing from CV and NLP, generative models have also achieved great success in many other areas like audio synthesis, control theory and various scientific discovery. Generally speaking, generative models aims to find the distribution of data, a probability density function defined on $\mathbb{R}^n$, then sample from it, whose output is i.i.d. Euclidean random variable with the input data.\par
%Common sense of physics tells us that the effects of spatial curvature cannot be ignored if we aim to establish the precise model. 
% Now, to deal with the geometric-prior data distributions, we must design Riemannian generative models for more 

To ensure appropriate representation learning and better generation performance on Riemannian manifolds, we developed a generative model named \emph{Riemannian Neural Geodesic Intepolant} in this work. The RNGI is a continuous-time stochastic process defined on a Riemannian manifold $\mathcal{M}$, connecting two probability distributions along the stochastic geodesics in the finite temporal interval $[0,1]$, then can generate from the endpoint distribution via efficient samplers. We summarize the main technical \textbf{contributions} based on the proposed RNGI as follows.  
 \begin{itemize}
    \item We extend the theoretical framework of stochastic interpolants to Riemannian manifolds, and prove that the temporal marginal probability density of RNGI solves a transport equation on Riemannian manifold under mild integrable condition of the velocity field. 
    \item By specifying the interpolating path as geodesics, we design a quadratic loss function for learning the velocity field of the RNGI and apply implicit score matching technique for learning the score field. We further provide theoretical bounds on the generative bias measured in terms of Wasserstein-2 distance between target and the neural approximated probability distributions in Proposition \ref{KL}. 
    \item We explore efficient sampling algorithms based on the learned vector field. We apply both Geodesic Random Walk method and Embedding SDE method to numerically simulate the stochastic differential equation corresponding to the transport equation, where the Embedding SDE method is proposed to reduce the accumulated error caused by the excessive discretization of Riemannian Brownian motion in the GRW method.
    \item We conduct experiments on benchmark datasets which are distributed on manifolds $\mathbb{S}^2$ and $\mathbf{SO}(3)$. Results show the flexible interpolating ability, higher likelihood compared to the baseline generative models and the improved generation performance with E-SDE method of RNGI.
\end{itemize}
 %We consider data generation process as a interpolation process between source random variable and target random variable on manifold, which can be quantitively modeled as a time-continuous stochastic process. 

%\subsection{Main Contribution}
%Our work builds a bridge between the flow-based generative models and SDE-based generative models on manifold. The Stochastic Intepolant framework is theoretically based on the transportation nature of flow, but, with the proved property that RNGI is identically distributed with SDE's solution process, also takes advantage of SDE's numerical approximation in the stage of sampling. We point out that, to our's best knowledge, this is the first implementation of such combination between the two fundamental generative models on manifold. Specifically, our contribution is two-fold: 

\subsection{Related Work}
We briefly review the related work on generative models on Riemannian manifolds. One type of work is to generalize the flow-based generative models to Riemannian manifold. The RCNF \citep{RCNF} and RNODE \citep{RNODE} generalize the two famous Euclidean flow-based model Normalizing Flow \citep{NMLZF} and Neural ODE \citep{NODE} to Riemannian setting. Both works build ODE flow path for sampling and train the model with log likelihood-related loss, relying on adjoint computation of projected ODE solver on $\mathcal{M}$. In \citep{NFTS}, the proposed method straightforwardly constructs normalizing flow on $\mathbb{S}^2$ and $\mathbb{T}^2$ by combining geometric operation blocks. \citep{MF,MCNF} formulate the flow process via transport equations and approximate the path of transportation by divergence-based loss functions. \citep{FMG} generalizes the Euclidean flow matching technique to Riemannian setting by defining the conditional flow as the integral curve of vector field, then matches the field via Riemannian distance on $\mathcal{M}$.

The other type of work constructs the diffusion-based generative models on Riemannian manifolds. The RSGM \citep{RSGM} model generalizes the score-based diffusion model \citep{SY}, deriving the forward-backward perturbing-sampling SDE and approximates the unknown heat kernel on manifold by both Sturm-Liouville decomposition and Varadhan's asymptotic. And concurrently \citep{RDM} establish the Riemannian diffusion model from the perspective of variational inference by optimizing the proposed Riemannian CT-ELBO, which is proven to be equivalent to the score-matching loss. \citep{RDB} generalizes diffusion bridges \citep{DB} to manifold setting, conditioning the destination of SDE at empirical distribution of training samples then matching the bridge model from both endpoints. As another genre, \citep{RNSDE} doesn't rely on the forward-backward scheme but directly consider parameterized Riemannian SDE, expressed in local coordinates, as universal transformer from the initial to the terminal distribution, and tunes the NN-surrogated drift and diffusion coefficient based on optimal transportation.

Different from these methods, our proposed RNGI model generalizes the stochastic interpolants to Riemannian manifolds, which can be regarded as a unifying framework for flows and diffusions. 

\section{Preliminaries}
\subsection{Stochastic Interpolant}

We begin by recalling the stochastic interpolant on Euclidean space \citep{SIs}. Let $\{ \Omega, \mathcal{F}, \mathbb{P}\}$ be a probability space, $x_0$ and $x_1$ be two  random variables from $\Omega$ to  $\mathbb{R}^d$. A stochastic interpolant between $x_0$ and $x_1$  is a stochastic process defined as:
\begin{equation}\label{GI}
    x_t = I(t;x_0,x_1),
\end{equation}
where $I\in C^2([0,1]; C^2(\mathbb{R}^d\times \mathbb{R}^d))$ denotes a differentiable bridge process such that
$$ I(0;x_0,x_1)=x_0, \quad I(1;x_0,x_1)=x_1. $$

The stochastic interpolant $x_t$ is a continuous-time stochastic process whose realizations are samples from the distribution of $x_0$ at $t=0$ and from the distribution of $x_1$ at $t=1$ by construction. Specifically, the initial distribution $x_0$ can, but not necessarily, be chosen as simple uniform distribution $U({\mathbb{R}^d})$ or Gaussian distribution $N(\mathbb{R}^d)$, while $x_1$, may be fairly complex, models the unknown distribution of the observational data. 
\subsection{Riemannian Manifold}
This paper considers complete, connected, smooth Riemannian manifolds $\mathcal{M}$ equipped with Riemannian metric $g$ as basic domain over which the generative model is learned. The tangent space of $\mathcal{M}$ at point $x$ is denoted by $T_x\mathcal{M}$ and $g$ defines an inner product over the tangent space denoted $\langle u,v\rangle_g, u,v\in T_x\mathcal{M}$. $T\mathcal{M}=\cup_{x\in\mathcal{M}}\{x\}\times T_x\mathcal{M}$ is the tangent bundle that collects all the tangent space of the manifold. The norm on Euclidean space and tangent space will be both denoted by $|\cdot|$. We use $\nabla$ and $\mathbf{div}$ to denote the gradient operator and divergence operator with respect to the spatial variable $x$ on $\mathcal{M}$, respectively.    Readers may refer to Appendix \ref{appendix:SDG} or \citep{Lee13} for a more comprehensive background on Riemannian manifolds. 
%\subsection{Notations}

Throughout this paper, $\mathcal{X}(\mathcal{M})$ is the space of vector fields on $\mathcal{M}$, $C(\mathcal{M})$ is the space of continuous functions on $\mathcal{M}$, $C^n(\mathcal{M})$ is the space of $n$-th continuously differentiable functions on $\mathcal{M}$. $C^n([0,1]; \mathcal{Y})$ denotes the space of $n$-th continuously differentiable functions from $[0,1]$ to some function space $\mathcal{Y}$ mentioned above. 

\section{Geodesic Interpolant on Riemannian manifold}

We first define \emph{Stochastic Interpolant} on Riemannian manifold by replacing Euclidean space with $\mathcal{M}$ : 
\begin{align*}
    x_t=I(t;x_0,x_1): [0,1]\times \mathcal{M}\times \mathcal{M} \rightarrow \mathcal{M}.
\end{align*}
For a generation task in practice, the initial distribution usually is chosen as a simple distribution on $\mathcal{M}$, e.g., uniform distribution $U(\mathcal{M})$ or Wrapped-Gaussian distribution $N(\mathcal{M})$ for sampling convenience. The marginal probability density $\rho(t,x)$ of the stochastic interpolant $x_t$ can bridge the initial probability density $\rho_0$ and final probability density $\rho_1$ corresponding to $x_0$ and $x_1$ respectively. Although connecting $x_0$ and $x_1$, the interpolant $I$ is unfeasible to directly be used for sampling since the distribution of $x_1$ is intractable. In the next section, we will establish ways for sampling through analyzing the evolution of the intermediate probability density $\rho(t,x)$.

%By construction, $x_t$ reaches the goal distribution $x_1$ precisely in finite time, but with the tightly coupled form of $I$, whose explicit expression is computationally intractable, it's unfeasible to directly sample from \ref{GI}.  To overcome this obstacle, we have shown the law of density evolution of $x_t$, i.e. the Riemannian transport equation.
%In this section, we will show that the density evolution of $x_t$ follows a transport equation on Riemannian manifold $\mathcal{M}$. 
Firstly, we define the velocity field of a stochastic process as follows. 

\begin{definition}[Velocity field]\label{VF}
Let $x_t$ be a stochastic process on Riemannian manifold $\mathcal{M}$ which is differentiable with respect to the time variable $t$. The velocity field $v(t,x)$ of $x_t$ is defined as:
\begin{equation}\label{vf}
    v(t,x) \triangleq \mathbb{E}(\partial_t x_t | x_t = x),
\end{equation}where $\partial_t x_t:\Omega\to T\mathcal{M}$ is a stochastic tangent field along the stochastic differentiable curve $x_t$.
\end{definition}
%We note that $\partial_t x_t:\Omega\to T\mathcal{M}$ is a stochastic tangent field along the stochastic differentiable curve $x_t$, which is slightly different from the instinct in Euclidean space. 
%Due to the non-flat nature of general Riemannian manifold, integral of vector field may rely on the parallel transport on $\mathcal{M}$ to perform linear operation between different tangent space. 
%But our Definition \ref{VF}, as a conditional expectation of $\partial_t x_t$ given $x_t = x$, is only local integral of $\partial_t x_t$ in a single tangent space $T_x\mathcal{M}$, where there is complete Euclidean structure to define vectorized integral.
Please note that in the above definition, the conditional expectation is taken by given $x_t=x$ and it can be locally calculated in a single tangent space $T_x\mathcal{M}$, which is the normal integral in the Euclidean space. With the well-defined velocity field, we now introduce a fundamental theorem of the evolution of the density $\rho(t,x)$, on which almost all of properties and algorithms of the model are based. 

\begin{theorem}\label{Thm}
    For Riemannian Stochastic Interpolant $x_t$ on manifold $\mathcal{M}$ with Riemannian metric $g$, we assume that its spatial-temporal probability density function $\rho(t,x)$ exists and $\rho(t,x)\in C^1([0,1]; C^2(\mathcal{M}))$. Denote the velocity field defined above by $v(t,x)$, and if $v(t,x)$ satisfies the integrable condition
    \begin{equation}\label{reg}
        \int_{\mathcal{M}}|v(t,x)|\rho(t,x)\mathrm{d}M(x) < \infty, 
    \end{equation}
    then $\rho(t,x)$ solves the transport equation on $\mathcal{M}$ as below:
    \begin{equation}\label{tsp}
        \partial_t \rho + \mathbf{div}[\rho \cdot v(t,x)] = 0.
    \end{equation}
    %where $\nabla\cdot$ is the divergence operator on $\mathcal{M}$.
\end{theorem}

Readers may refer to the supplementary materials for complete proof of Theorem \ref{Thm} and the detailed verification of \ref{reg}. Next, we generalize the score function in Euclidean case to the score field on $\mathcal{M}$, which is a gradient field generated by scalar function $\rho(t,x)$.
\begin{definition}[Score field]\label{SF}
Let $x_t$ be a stochastic process living on Riemannian manifold $\mathcal{M}$, and its score field $s(t,x)$ is defined as:
\begin{equation}
    s(t,x) \triangleq \nabla \log \rho(t,x).
\end{equation}
%where $\nabla$ is the gradient operator on $\mathcal{M}$.
\end{definition}

Consider the perturbation of velocity field as
\begin{equation}\label{ptb}
    v_F(t,x) = v(t,x)+\epsilon(t)s(t,x),
\end{equation}
then the transport equation (\ref{tsp}) can be transformed into the following Fokker-Planck Equation which is also solved by $\rho$
\begin{equation}\label{FPE}
    \partial_t \rho + \mathbf{div} (\rho\cdot v_F) = \epsilon(t)(\Delta_{\mathcal{M}})\rho,
\end{equation}
where $\epsilon(t)\geq 0$ is a coefficient and  $\Delta_\mathcal{M}$ denotes the Laplace-Beltrami operator on $\mathcal{M}$. Parallel with classical stochastic analysis theory, Fokker-Planck equation (\ref{FPE}) on $\mathcal{M}$ also corresponds to a stochastic differential equation on $\mathcal{M}$. More precisely, the solution $\rho(t,x)$ to \ref{FPE} is concurrently the time-marginal probability density of $\overrightarrow{X}_t$ that follows the stochastic differential equation as:
\begin{align}\label{SDE}
    \mathrm{d}\overrightarrow{X}_t = v_F(t,\overrightarrow{X}_t)\mathrm{d}t&+\sqrt{2\epsilon(t)}\mathrm{d}B_t^{\mathcal{M}},\\
    \overrightarrow{X}_t|_{t=0}& = x_0,\notag
\end{align}
where $B_t^{\mathcal{M}}$ is the Brownian motion on $\mathcal{M}$. And for the backward perturbation and backward Fokker-Planck equation
\begin{align*}
   & v_B(t,x) = v(t,x)-\epsilon(t)s(t,x),\\
   & \partial_t \rho + \mathbf{div} (\rho\cdot v_B) = -\epsilon(t)(\Delta_{\mathcal{M}})\rho,
\end{align*}
there is an associated backward SDE
\begin{align}\label{BSDE}
    \mathrm{d}\overleftarrow{X}_t = -v_B(1-t,\overleftarrow{X}_t&)\mathrm{d}t+\sqrt{2\epsilon(t)}\mathrm{d}B_t^{\mathcal{M}},\\
    \overleftarrow{X}_t|_{t=0}& = x_1\notag
\end{align}
whose solution $\overleftarrow{Y}_t = \overrightarrow{X}_{1-t}$. As there is no essential difference between \eqref{SDE} and \eqref{BSDE}, so we mainly discuss \eqref{SDE} as representative in following sections. And when we set $\epsilon(t)=0$, both of  \eqref{SDE} and \eqref{BSDE} degenerate to an ODE
%Concurrently, the correlation between the governing Fokker-Planck equation \ref{FPE} and transport equation \ref{tsp} also reveals the correlation between sampling SDE \ref{SDE} and below probability flow ODE:
\begin{equation}\label{ODE}
    \mathrm{d}\overleftrightarrow{X}_t = v_F(\overleftrightarrow{X}_t,t)\mathrm{d}t,
\end{equation}
which can be bi-directionally solved between $t=0$ and $t=1$.
%which, as a special case of \ref{SDE} when $\epsilon=0$, shares the same time-marginal density $\rho(t,x)$ with $X_t^{SDE}$.
%and generates a continuous flow $\phi(t,x)$ on $\mathcal{M}$.

The induced SDE (\ref{SDE}, \ref{BSDE}) and ODE (\ref{ODE}) establish surrogate models $X_t$, equivalent to stochastic interpolant $x_t$ in the sense of distribution and they can be used for simulation by leveraging SDE or ODE solvers on Riemannian manifold.

Till now the information of unknown data distribution $\rho_1$ is fully carried by the drift vector field $v_F$, which can be learned from data. We will introduce the learning and sampling algorithms in Section \ref{Sec:Alg}. Before that, we introduce the \emph{Geodesic Interpolant} and its explicit form on example manifolds. %and then we can approximately sample from $\rho_1$ via the explicit computation of $X_1$.

%In fact, $x_t$ is not only a Riemannian stochastic process, but also a stochastic path on $\mathcal{M}$ connecting two stochastic points $x_0$ and $x_1$, and different $t\in [0,1]$ of $x_t$ denotes the different stochastic intermediate interpolation point between $x_0$ and $x_1$. 
As the velocity field relies on the partial derivative of the interpolating path $I(t;x_0,x_1)$, we need to specify the form of $I$ for constructing the learning algorithm of the velocity for computational tractable. In this section, we specify the stochastic interpolant $I(t;x_0,x_1)$ on $\mathcal{M}$ as the geodesic, the shortest path connecting two random points on manifold. %like the straight line do in Euclidean space.

\begin{definition}[Riemannian Geodesic Interpolant]\label{Geo}
    Given a probability space $\{\Omega, \mathcal{F},\mathbb{P}\}$ and two random variables $x_0, x_1: \Omega\rightarrow\mathcal{M}$, the Geodesic Interpolant $I(t;x_0,x_1)$ between $x_0$ and $x_1$ is defined as
    \begin{equation}\label{GI}
        I(t;x_0,x_1) = \mathrm{Exp}_{x_0} (t \cdot \mathrm{Log}_{x_0}x_1),
    \end{equation}
    where $\mathrm{Exp}_{x_0}$ and $\mathrm{Log}_{x_0}$ denote the exponential and logarithm map at point $x_0$ respectively.
\end{definition}

To construct the geodesic interpolant rigorously, we must ensure that the geodesics are 'uniquely reachable' on (almost) the entire $\mathcal{M}$. But generally speaking, the exponential and logarithm map are usually only defined in a small neighborhood of $x_0$. As stochastic points on $\mathcal{M}$, $x_0$ and $x_1$ are not assumed to be closed with each other. Therefore, we must extend the locally defined exponential and logarithm map to almost the entire manifold, except for at most the cut locus of zero measure, which requires the manifold is geodesically complete\citep{Lee13}.

\begin{definition}[Geodesic completeness]
    Let $\mathcal{M}$ be a connected manifold equipped with Riemannian metric $g$. If for any geodesic $\gamma_t$ on $\mathcal{M}$, the domain of $\gamma$ can be extended from $[0,1]$ to $\mathbb{R}^+$, then $\mathcal{M}$ is called as geodesically complete.
\end{definition}

The following theorem and its corollaries guarantee the well-posedness of the geodesic interpolant \ref{Geo} even for 'distant' random points $x_0$ and $x_1$ on our concerned manifolds.

\begin{proposition}[\citep{HopfRinow}]
    If $\mathcal{M}$ is a connected manifold equipped with Riemannian metric $g$, and $d_g(\cdot,\cdot)$ is the Riemannian distance on $\mathcal{M}$ induced by $g$, then 
    \begin{equation*}
        \mathcal{M}\ \text{is geodesically complete} \Leftrightarrow (\mathcal{M},d) \ \text{is complete metric space}.
    \end{equation*}
\end{proposition}
\begin{corollary}
    (1) If $\mathcal{M}$ is a compact manifold, then $\mathcal{M}$ is geodesically complete. (2) If $\mathcal{M}$ is a compact and complete manifold, the there exists a geodesic on $\mathcal{M}$ joining any two points $p,q$ on $\mathcal{M}$.
\end{corollary}

Now for a wide range of Riemannian manifolds, especially the compact ones like hypersphere, the geodesics are 'reachable' between the initial random variable $x_0$ and the target random variable $x_1$ on them. We still to make geodesics unique by discard a few singular points, i.e. the cut locus.

\begin{definition}[Cut locus]
On a Riemannian manifold $\mathcal{M}$, a point $q$ is called cut point to the point $p$ if there are two or more minimizing geodesics joining $p$ and $q$. The set of all such $q$, denote by $\mathcal{C}(p)$ is defined as the cut locus of $p$.
\end{definition}

Actually, $\mathcal{C}(p)$ is the collection of the points which make the exponential map $\mathrm{Exp}_p$ from $p$ not injective, so the logarithm map $\mathrm{Log}[\mathcal{C}(p)]$ cannot be defined. As for constructing geodesic interpolant, we must exclude $\{x_1(\omega):x_1(\omega)\in \mathcal{C}[x_0(\omega)]\}$ from the target dataset, and project them back into $\mathcal{M}-\mathcal{C}(x_0)$ to keep the information contained in them as much as possible.

\section{Neural Geodesic Interpolant Algorithm}\label{Sec:Alg}

To parameterize the geodesic interpolant model \ref{GI} and improve its generalization ability, we use neural networks as the approximators of latent vector fields on Riemannian manifold, which leads to a NN-based surrogate model - Riemannian Neural Geodesic Interpolant (RNGI).

\subsection{Network training}
The construction of sampling SDE \ref{SDE} is based on the perturbation in Equation \ref{ptb}, which consists two unknown vector fields on $\mathcal{M}$ to be learned. We use two neural networks to approximate velocity field $v(t,x)$ and the score field $s(t,x)$ and denote the approximate value as $\hat{v}$ and $\hat{s}$ respectively. For the learning of velocity field, we minimize the following quadratic loss function $\mathcal{L}_v[\hat{v}]$:
\begin{equation}\label{qrl}
    \int_0^1\mathbb{E}_{x_t} \left[ \frac{1}{2}|\hat{v}(x_t,t)|_g^2 - \langle\partial_tI(t;x_0,x_1) , \hat{v}(x_t,t)\rangle_g \right]\mathrm{d}t.
\end{equation}
From the properties of quadratic function, the vector field defined in Definition \ref{VF} is the unique minimizer of $\mathcal{L}_v$ in $\mathcal{X}(\mathcal{M})$. And for the learning of the score field, we optimize the implicit score-matching (ISM) loss $\mathcal{L}_s[\hat{s}]$: 
\begin{equation}\label{ism}
     \int_0^1 \mathbb{E}_{x_t} \left[\frac{1}{2}|\hat{s}(x_t,t)|_g^2 + \mathbf{div}\hat{s}(x_t,t)\right] \mathrm{d}t,
\end{equation}
then the vector field defined in Definition \ref{SF} is the unique minimizer of $\mathcal{L}_s$ in $\mathcal{X}(\mathcal{M})$, which has been proven in \citep{JMLR}. Note that unlike denoising score-matching (DSM) loss in Diffusion Models, ISM allows for score approximation of general stochastic process $x_t$, without any structural constraint like Brownian motion or Ornstein-Uhlenbeck process.

Both the finding of the optimal neural network solution of \ref{qrl} and \ref{ism} make up the neural approximation of geodesic interpolant, and the detailed training process is shown in Algorithm \ref{Tvs}.

\begin{algorithm}
    \caption{Training of velocity field $v(t,x)$ and score field $s(t,x)$}
    \label{Tvs}
    \textbf{Input:} \textit{Prior samples} $\{x_0^{(n)}\}$, \textit{Data samples} $\{x_1^{(n)}\}$, \textit{Interpolating path} $I(t;x_0,x_1)$, \textit{Initialized velocity field network} $v_{\theta_0} (t,x)$, \textit{Initialized score field network} $s_{\eta_0} (t,x)$, \textit{Epoch number} $M$,  \textit{Batch size} $N$
    
    \begin{algorithmic}[1]
    \For{$j = 1,\dots, M$}
    \State  Sample $\{t^{(i)}\} \sim \textrm{Uniform}[0,1]$
    \State  Draw a batch of data $\{x_0^{(i)}\}$ and $\{x_0^{(i)}\}$
    \For{$i = 1, \dots, N$}
    \State  Create coupled triples $[t^i,x_0^i,x_1^i]$
    \State  Calculate interpolation point $x_t^i = I(t^i;x_0^i,x_1^i)$ and velocity $\partial_t I(t^i;x_0^i,x_1^i)$
    \State  Calculate loss function $\mathcal{L}_v(v_\theta)$ and take gradient descent step on hyperparameter $\theta$
    \State  Calculate loss function $\mathcal{L}_s(s_\eta)$ and take gradient descent step on hyperparameter $\eta$
    \EndFor
    \EndFor
    \State \textbf{Return} Neural network $v_{\theta^*}$ and $s_{\eta^*}$ with the optimal parameter $\theta^*$ and $\eta^*$
    \end{algorithmic}
\end{algorithm}

\subsection{Samples generation}

The marginal probability density of $x_t$ is equal to that of $X_t^{SDE}$ defined in Equation (\ref{SDE}) at any time $t$, so sampling from the terminal distribution of $x_1$ with abstract form $I$ can be realized by sampling from $X_t^{SDE}$, which further can be approximately implemented by numerically solving the neural parameterized stochastic differential equation on $[0,1]$:
\begin{equation}\label{pSDE}
    \mathrm{d}\widehat{X}_t = v_F^{\theta^*,\eta^*}(\widehat{X}_t,t)\mathrm{d}t+\sqrt{2\epsilon(t)}\mathrm{d}B_t^{\mathcal{M}}, \quad \widehat{X}_0=x_0.
\end{equation}
\subsubsection{Geodesic Random Walk}
In the context of Riemannian generative modeling, Geodesic Random Walk (GRW), as an intrinsic approach to solving Riemannian SDE without any dependencies on the ambient space, is a widely-used discretization scheme. GRW's step is based on the sampling on tangent bundles $T\mathcal{M}$ and projecting the sampled tangent vector to ambient space $\mathcal{M}$:
\begin{equation}\label{GRW}
    \widehat{X}_{t+1} = \mathrm{Exp}_{\widehat{X}_t} (v_F^{\theta^*,\eta^*}(\widehat{X}_t,t)\Delta t + \sqrt{2\epsilon\Delta t} \cdot Z_t),
\end{equation}
where $\Delta t$ is time step and $Z_t$ is a Gaussian random variable in $T_{\widehat{X}_t}\mathcal{M}$, and $\sqrt{\Delta t}$ comes from the second-order approximation of Riemannian Brownian motion $B_t^{\mathcal{M}}$. 

It's provable that GRW defined as \ref{GRW} will converge to the exact solution of \ref{pSDE} in the sense of semi-groups \citep{GRW1975}. Based on the convergence result, intuitively, GRW implements a numerical scheme to model the stochastic dynamics governed by Riemannian SDE like its Euclidean counterpart. In the following context, when we mention ODE method for sampling, we mean that $\epsilon=0$ in Equation (\ref{pSDE}).

\subsubsection{Embedding SDE}

Despite being a classical method for SDE discretization on $\mathcal{M}$, the GRW scheme may produce biased samples due to the intrinsic approximation of Brownian motion. When $\mathcal{M}$ is compact, the exponential map used in GRW can only be bijective in a bounded  domain $D\in T_{\widehat{X}_t}\mathcal{M}$. But $Z_t$, as an unbounded Gaussian random variable in $T_{\widehat{X}_t}\mathcal{M}$ to simulate the Brownian part in \ref{pSDE}, still has probability to escape from $D$. Then the overstepping tangent vector will be $\mathrm{Exp}$-mapped to a circular point on $\mathcal{M}$, generating unexpected image point and being iterated into wrong sample after multi-step updates.

So in brief, the insufficient expressivity of local coordinates injects systematic error into the GRW simulation of \ref{pSDE}, preventing the stochastic generation scheme of SI models from competing with the deterministic scheme. Fortunately, stochastic analysis theory opens a broader view to deal with the simulation of SDE on manifold, establishing a new way to fix the locality issue. 
 
From the perspective of topology, both of classical Whitney's\cite{Whitney1936} and Nash's theorem \citep{Nash54} state that a $m$-dimensional Riemannian manifold can always be embedded in a Euclidean space with higher dimension $d\geq m+1$, then extrinsically parameterized by the global frame in $\mathbb{R}^d$ as
\begin{equation*}
    \mathcal{M} = \{p|p=(x_1,\dots,x_d)\in\mathbb{R}^d\}.
\end{equation*}
Based on such embedding, Proposition \ref{BM} gives another construction of Brownian motion on manifold.

\begin{proposition}[\citep{PeiHsu}]\label{BM}
    If $\mathcal{M}$ is a sub-manifold of $\mathbb{R}^d$, then we can construct the Brownian Motion $B_t^{\mathcal{M}}$ on $\mathcal{M}$ by projecting the $d-$dimensional Euclidean Brownian Motion $W_t$ in $\mathbb{R}^d$ onto $\mathcal{M}$:
    \begin{equation}\label{proj}
        \mathrm{d}B_t^{\mathcal{M}} = P_\alpha(B_t^{\mathcal{M}})\circ \mathrm{d}W_t^\alpha,
    \end{equation}
    where $P_\alpha(\cdot)$ denotes the $\alpha-$th orthonormal projection operator from $\mathbb{R}^d$ to the tangent space $T_{(\cdot)}\mathcal{M}$, '$\circ \mathrm{d}W_t$' denotes the Stratonovich stochastic differential and the index $\alpha$ follows the Einstein summation convention.
\end{proposition}

Proposition \ref{BM} generates $\mathcal{M}$-Brownian motion $B_t^{\mathcal{M}}$ by solving Stratonovich SDE \ref{proj} in ambient Euclidean space $\mathbb{R}^d$ without local Gaussian sampling, avoiding the accumulation of errors due to the incomplete expressiveness of the local coordinate used by GRW. Having composed the embedding diffusion term $\mathrm{d}B_t^{\mathcal{M}}$ and learned drift term $v_F^{\theta^*,\eta^*}: \mathcal{M} \to T\mathcal{M}$, we complete the Embedding SDE (E-SDE) scheme of \ref{pSDE} as

\begin{equation}\label{eSDE}
    \mathrm{d}X_t^{SDE} = v_F^{\theta^*,\eta^*}(t,X_t^{SDE})\mathrm{d}t+\sqrt{2\epsilon(t)} P(X_t^{SDE})\circ \mathrm{d}W_t.
\end{equation}

Contrast to the conventional GRW scheme for stochastic generation, our proposed E-SDE scheme not only achieves higher accuracy by globally simulating Riemannian Brownian motion $B_t^{\mathcal{M}}$ instead of locally sampling $Z_t\sqrt{\Delta t}\approx \mathrm{d}B_t^{\mathcal{M}}$, but also enjoys faster convergence with the help of the developed numerical solvers of \ref{eSDE} in $\mathbb{R}^d$. \citep{cheng2022} has proven that the convergence order of GRW is $\mathcal{O}(\sqrt{\Delta t})$ as the stepsize $\Delta t\to 0$, and for the convergence order of various discretization schemes\citep{NAS2011} of E-SDE, readers may refer to Table \ref{tab:NSC}.

\begin{table}[htbp]
    \caption{The relation between the error and discretization stepsize of numerical schemes for SDE on Riemannian manifolds.}
    \label{tab:NSC}
    \centering
    \vspace{5pt}
    \begin{tabular}{lr}
    \toprule
    schemes & orders\\
    \midrule
    GRW: Geometric Euler-Maruyama & $\mathcal{O}(\sqrt{\Delta t})$ \\
    \midrule
    E-SDE: Euler-Maruyama & $\mathcal{O}(\sqrt{\Delta t})$\\
    E-SDE: Euler-Heun& $\mathcal{O}(\sqrt{\Delta t})$\\
    E-SDE: Milstein & $\mathcal{O}(\Delta t)$\\
    E-SDE: Stochasitc Runge-Kutta & $\mathcal{O}[(\Delta t)^{1.5}]$\\
    \bottomrule
\end{tabular}
\end{table}

We summarize the entire sampling process in Algorithm \ref{Smp} below.

\begin{algorithm}
    \caption{Sampling of new data from objective distribution $x_1$}
    \label{Smp}
    \textbf{Input:} \textit{velocity field network} $v_{\theta^*}$, \textit{score field network} $s_{\eta^*}$, \textit{number of steps} $N$, \textit{diffusion coefficient } $\epsilon (t)$, \textit{number of samples} $M$, \textit{prior distribution} $x_0\sim \rho_0$
    
    \begin{algorithmic}[1]
    \State Calculate the perturbed drift: $v_F^{\theta^*,\eta^*} = v_{\theta^*} + \epsilon (t)s_{\eta^*}$ and time interval $\triangle t = \frac{1}{N}$
    \For{$i = 1,\dots, M$}
    \State Draw new sample $x_0^{i}$ from prior distribution $\rho_0$
    \For{$t=0,\Delta t, \dots, N\Delta t=1$}
    \State Numerical simulate SDE: $\mathrm{d}X_t = v_F^{\theta^*,\eta^*}\mathrm{d}t + \sqrt{2\epsilon (t)} \mathrm{d}B_t^{\mathcal{M}}$ with initial value $x_0^{i}$ by GRW or ESDE
    \EndFor
    \EndFor
    \State \textbf{Return} New samples $\{x_1^{i}\}_{i=0}^M$ from data distribution $\rho_1$
    \end{algorithmic}       
\end{algorithm}

\subsection{Error estimation}

For the parameterized surrogate generating process $\widehat{X}_t$, there is always unavoidably some systematic error due to the suboptimal neural network approximation of $v_F(t,x)$. As both the interpolating process and generating process be modeled as SDE, we can bound the gap between the target distribution and the sampled distribution in the sense of Wasserstein-2 distance.

\begin{proposition}\label{KL}
    For two Riemannian stochastic process $X_t$ and $\widehat{X}_t$ living on $\mathcal{M}$, whose time marginal probability density function are denoted by $\rho(t,x)$ and $\hat{\rho}(t,x)$ and corresponding velocity field denoted by $v(t,x)$ and $\hat{v}(t,x)$ respectively.
    If both of their density evolution satisfy transport equation, i.e.
    \begin{align*}
       &\partial_t \rho + \mathbf{div}[\rho \cdot v(t,x)] = 0, \\\label{AKL2}
       &\partial_t \hat{\rho} + \mathbf{div}[\hat{\rho} \cdot \hat{v}(t,x)] = 0.
    \end{align*}
    Then the KL divergence between the target distribution $\rho_1$ and generated distribution $\hat{\rho}_1$ can be integrated as:
    \begin{equation}\label{KLe}
        \mathrm{KL}(\rho_1||\hat{\rho}_1) = \int_0^1 \mathrm{d}t\int_\mathcal{M} \langle s-\hat{s},v-\hat{v} \rangle_g\rho(t,x)\mathrm{d}M(x).
    \end{equation}
\end{proposition}
Then from \ref{KLe}, the generative discrepancy can be jointly controlled by the approximation error of $v$ and $s$.

\begin{proposition}\label{WP}
Consider two SDEs with same initial data $x_0$ and Brownian motion $B_t^{\mathcal{M}}$ on $\mathcal{M}$:
\begin{align*}
\mathrm{d}X_t &= v_F(X_t,t)\mathrm{d}t+\sqrt{2\epsilon(t)}\mathrm{d}B_t^{\mathcal{M}},\\
\mathrm{d}\hat{X}_t&= \hat{v}_F(\hat{X}_t,t)\mathrm{d}t+\sqrt{2\epsilon(t)}\mathrm{d}B_t^{\mathcal{M}}.
\end{align*}
Let $\mu_t$ and $\hat{\mu}_t$ be  the probability measures induced by $X_t$ and $\hat{X}_t$ respectively, assume that $v$ and $\hat{v}$ both are Lipschitz with constant $L$, we have
\begin{equation}\label{eq:WP}
        W_p(\mu_t, \hat{\mu}_t) \leq e^{Lt}
       \sup_{s\in [0,t]}\sup_{x\in\mathcal{M}}|v_F(x,s) - \hat{v}_F(x,s)|_g, 
    \end{equation}
    where $W_p$ denotes the Wasserstein $p-$distance for $p\geq 1$. Let $t=1$ in \ref{eq:WP} then we have the final error estimation of $\hat{\mu}_1$.
\end{proposition}

Both of the proof of proposition \ref{KL} and \ref{WP} are collected in the Appendix \ref{appendix:prf}.

\subsection{Interpolant construction}

Next, we show the analytic form of the geodesic interpolant and construct E-SDE scheme on two important kinds of Riemannian manifolds: Hypersphere $\mathbb{S}^n$ and (matrix) Lie group $\mathbf{SO}(3)$ for their broad-application in geometric data science. And the detailed completeness analysis of $\mathbb{S}^n$ and \textbf{SO}(3) are in supplementary materials.

\paragraph{Hypersphere $\mathbb{S}^n$} The hypersphere $\mathbb{S}^n$ can be naturally embedded in the Euclidean space $\mathbb{R}^{n+1}$, and the exponential map $\mathrm{Exp}_{(\cdot)}$ on $\mathbb{S}^n$ can be written as
\begin{equation}
    \mathrm{Exp}_p (v) = \cos(|v|)p + \sin(|v|)\frac{v}{|v|},
\end{equation}
where $p$ is a point on $\mathbb{S}^n$ and $v$ is a tangent vector in tangent space $T_p \mathbb{S}^n$ at $p$. Similarly the logarithm map $\mathrm{Log}_{(\cdot)}$ can be written as:
\begin{equation}
    \mathrm{Log}_p (q) = \frac{\arccos \langle p,q\rangle}{|q-\langle p,q\rangle p|} [q-\langle p,q\rangle p],
\end{equation}
where $p$ and $q$ both denote points on $\mathbb{S}^n$ and $\langle\cdot,\cdot\rangle$ denotes the standard inner product in $\mathbb{R}^{n+1}$.

Thanks to geodesic completeness of $\mathbb{S}^n$, the locally defined exponential and logarithm map can be extended to the entire $\mathbb{S}^n-\{-p\}$ only by continuously lengthening $v$ from $|v|=1$ to $|v|\to\pi$, and for the sparsity of a single point $-p$ in $\mathbb{S}^n$, we do not need to exclude any point from the target dataset. Therefore, the interpolating process $I(t;x_0,x_1)$ on $\mathbb{S}^n$ can be written as:
\begin{align}\label{GIS}
    I(t;x_0,x_1) &= \mathrm{Exp}_{x_0} (t \cdot \mathrm{Log}_{x_0}x_1)\notag \\
                 &= \cos(t \cdot |\mathrm{Log}_{x_0}x_1|)x_0\notag \\
                 &\quad+ \sin(t \cdot |\mathrm{Log}_{x_0}x_1|)\frac{\mathrm{Log}_{x_0}x_1}{|\mathrm{Log}_{x_0}x_1|}.
\end{align}

Now we consturct E-SDE scheme on $\mathbb{S}^n$. For any fixed point $x=(x_1,\dots,x_{n+1})$ on $\mathbb{S}^n \subset \mathbb{R}^{n+1}$ and any separated point $\xi\in\mathbb{R}^{n+1}$, the projection operator $P_x$ from $\mathbb{R}^{n+1}$ to the tangent space $T_x\mathbb{S}^n$ at $x$ is given by:
\begin{equation*}
    P_x(\xi) = \xi- \langle\xi,x\rangle x,
\end{equation*}
where $\langle\cdot,\cdot\rangle$ denotes the standard inner product in $\mathbb{R}^{n+1}$. So from above $P$'s expression in the form of geometric operator, we can also derive the matrix form of $P$ as $(P(x)_{ij})$ by entries:
\begin{equation*}
    P(x)_{ij} = \delta_{ij}-x_ix_j.
\end{equation*}
where $\delta$ denotes Kronecker symbol. And substituting $P(x)_{ij}$ into \ref{proj} gives a SDE in $\mathbb{R}^{d+1}$
\begin{equation*}\label{SSDE}
    \mathrm{d}B_t^i = \sum_{j=1}^{n+1} (\delta_{ij}-B_t^iB_t^j)\circ \mathrm{d}W_t^j,
\end{equation*}
where $B_t^i$ is the $i-$th component of $B_t$. The solution process $B_t$, also known as Stroock’s representation\citep{Stroock}, is a Brownian motion living on the hypersphere $\mathbb{S}^n$.

\paragraph{Lie group: \textbf{SO}(3)} 
The Lie algebra $\mathfrak{so}(3)$ of \textbf{SO}(3) is the space of skew-symmetric matrices, whose matrix element $\omega$ can also be expressed as a vector $\hat{\omega}=[\omega_1,\omega_2,\omega_3]$. The Rodrigues' rotation formula \citep{MSL94AMI} bridges 3D rotation vector and 3D orthogonal matrix, which gives the exponential map from $\mathfrak{so}(3)$ to \textbf{SO}(3):
\begin{equation}\label{Exp}
    \mathrm{Exp}_e(\omega) = I + \sin\theta \cdot \omega + (1-\cos\theta) \cdot \omega^2,
\end{equation}
where $\theta = \sqrt{\omega_1^2+\omega_2^2+\omega_3^2}$ denotes the Frobenius norm of $\omega$, $e$ denotes the identity element in the group. Inversely, the logarithm map from \textbf{SO(3)} to $\mathfrak{so}(3)$ at $e$ is given by:
\begin{equation}\label{Log}
    \mathrm{Log}_e
    \left[
    \begin{array}{ccc}
    R_{11} & R_{12} & R_{13} \\
    R_{21} & R_{22} & R_{23} \\
    R_{31} & R_{32} & R_{33}
    \end{array}
    \right]
    = \frac{\gamma}{2\sin\gamma}
    \left[
    \begin{array}{c}
    R_{32}-R_{23}  \\
    R_{13}-R_{31}  \\
    R_{21}-R_{12}
    \end{array}
    \right],
\end{equation}
where the rotation matrix $R\in\mathbf{SO}(3)$ is expressed by entries and rotation angle $\gamma$ is calculated by
\begin{equation*}
    \gamma = \arccos(\frac{\mathrm{tr}(R)-1}{2}).
\end{equation*}
For any point $p\in$\textbf{SO}(3), the $\mathrm{Exp}$ and $\mathrm{Log}$ map at $p$ are translated from those at $e$:
\begin{align*}
    &\mathrm{Exp}_p(v) = p\cdot \mathrm{Exp}_e(p^{-1}v),\\
    &\mathrm{Log}_p(q) = p\cdot\mathrm{Log}_e(p^{-1}q),
\end{align*}
where $v\in T_p\mathbf{SO}(3)$ and $q\in\mathbf{SO}(3)$. For the random variable $x_0$ and $x_1$, the geodesic interpolant $I(t;x_0,x_1)$ on \textbf{SO}(3) is expressed as
\begin{align}\label{GIL}
    I(t;x_0,x_1) &= \mathrm{Exp}_{x_0} (t\cdot \mathrm{Log}_{x_0}x_1)\notag \\
                 &= x_0\cdot \mathrm{Exp}_e [t \cdot \mathrm{Log}_e(x_0^{-1}x_1)].
\end{align}
The compactness of \textbf{SO}(3) makes it geodesically complete, and to ensure the uniqueness, note that the orthogonality of \textbf{SO}(3)'s matrix elements has a constraint on their trace that $\mathrm{tr}(R)\in[-1,3]$, so choose the main value of $\arccos$ then we get a mapped $\gamma\in[0,\pi]$. But for those matrices with $\mathrm{tr}=-1$, which represents a rotation with $\mathrm{angle}=\pi$, the bijection between rotation vectors and orthogonal matrices does not hold anymore, leading to {\tt nan} value of \ref{Log}. So for those datapoints $(x_0^{-1}x_1)(\omega)$ satisfying $\mathrm{tr}[(x_0^{-1}x_1)(\omega)] =-1$, we must truncate their trace when directly construct geodesic interpolant on $\mathbf{SO}(3)$.

To construct E-SDE scheme on $\mathbf{SO}(3)$, as the embedding of $\mathbf{SO}(3)$ into $\mathbb{R}^d$ is not as trivial as that of $\mathbb{S}^n$, we introduce the 'truncation-orthogonalization' embedding of $\mathbf{SO}(3)$ into $\mathbb{R}^6$ proposed by \citep{R62019}.

For a 3-d orthogonal matrix $R=(\mathbf{r}_1,\mathbf{r}_2,\mathbf{r}_3)$, the truncating map $\mathcal{P}:\mathbf{SO}(3)\to \mathbb{R}^6$ is defined by
\begin{equation*}
    \mathcal{P}(R) = \left(
    \begin{array}{c}
    \mathbf{r}_1 \\
    \mathbf{r}_2
    \end{array}
    \right).
\end{equation*}
And for a vector $\mathbf{l}=\left(\begin{array}{c}\mathbf{l}_1 \\\mathbf{l}_2\end{array}\right) \in \mathbb{R}^6$, the Gram-Schimit orgonalization of the two 3-d vectors $\mathbf{l}_1$ and $\mathbf{l}_2$ gives two orthogonal 3-d unit vectors $\mathbf{r}_1$ and $\mathbf{r}_2$. Additionally, the cross product gives another unit vector $\mathbf{r}_3 = \mathbf{r}_1 \times \mathbf{r}_2$, which is orthogonal to both $\mathbf{r}_1$ and $\mathbf{r}_2$. Then the orthogonalizing map $\mathcal{Q}:\mathbb{R}^6\to \mathbf{SO}(3)$ is defined by

\begin{equation*}
    \mathcal{Q}(\mathbf{l}) = (\mathbf{r}_1,\mathbf{r}_2,\mathbf{r}_3).
\end{equation*}

$\mathcal{P}$ and $\mathcal{Q}$ establish a bijection between $\mathbf{SO}(3)$ and a submanifold $\mathcal{M}$ of $\mathbb{R}^6$. Such bijection cannot be implemented by the Euler angle embedding, axis-angle embedding or quaternion embedding. Specifically, $\mathcal{M}$ can be further embedded as a submanifold of $\mathbb{S}^5$ in $\mathbb{R}^6$, so the probability distribution on $\mathbf{SO}(3)$ can be transformed to be supported on $\mathcal{M}\subset \mathbb{S}^5$, which enables us to construct RNGI model and its E-SDE scheme in $\mathbb{R}^6$.

\section{Experiment}
We apply the RNGI model to datasets living on two Riemannian manifolds: $\mathbb{S}^2$ and \textbf{SO}(3) to validate its effectiveness. Both of the geographical and synthetic density generation experiment aim to demonstrate the improved accuracy of RNGI models in terms of log-likelihood and E-SDE in terms of distance metric, and the distribution connection experiment aims to demonstrate the flexibility of RNGI models to generate samples of target distribution from arbitrary initial distributions. All of experiments run on a single Nvidia Geforce RTX 4090/3090 GPU.

\subsection{Geographical data on earth}

Many geographical and climatic events, such as volcanic eruptions \citep{Volcano}, earthquakes \citep{Earthquake}, floods \citep{Flood}, and wild fires \citep{Fire}, whose occurrence can be represented as geographical coordinates on the surface of the earth, can be naturally modeled as samples drawn from latent probability distributions $\rho_{event}$ on $\mathbb{S}^2$. We use the RNGI model to bridge the uniform distribution $U(\mathbb{S}^2)$ and the unknown $\rho_{event}$. Then we transform new samples from $U(\mathbb{S}^2)$ into high-quality samples from $\rho_{event}$ by the generative schemes of RNGI model.

\begin{figure}[htbp]
    \centering
    \includegraphics[width=\textwidth]{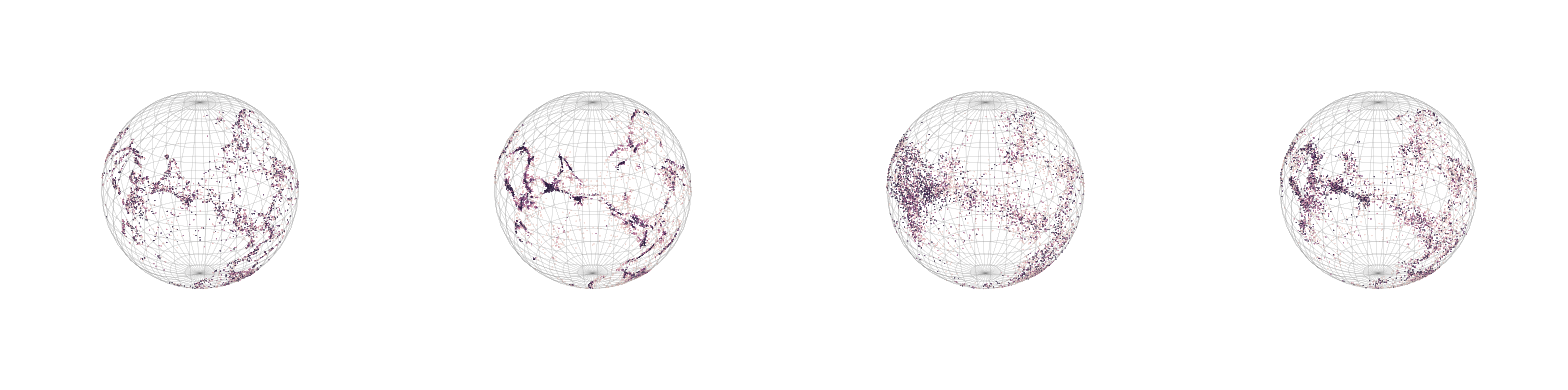}
    \caption{Generative visualization on flood dataset. From left to right there are the ground truth data, samples from ODE, GRW and E-SDE. The diffusion coefficient of GRW and E-SDE is set to $\epsilon=0.01$ and the sampling steps are $N=100$.}
    \label{Flood36}
\end{figure}

We use two fully-connected neural network to respectively predict the velocity field and score field of the geodesic interpolant on $\mathbb{S}^2$, and both of the inputs and outputs are extrinsic 3-d vector representation of spherical points. We employ 'RiemannianAdam' optimizer in package {\tt geoopt}\citep{Geoopt} on volcano, earthquake and flood and AdamW on fire datasets after tuning, and train each experiment for 60000 iterations on the first three experiments and 120000 iterations on the last one. The detailed parameter settings are listed in Table \ref{paras2} in the appendix \ref{subappendix:para}. The stochastic sampling via E-SDE is conducted by Euler-Heun solver of 0.5 order or Heun solver of 1.5 order in package {\tt torchsde}\citep{NSDE}. And the deterministic sampling via ODE is conducted by Euler or dopri5 solver in package {\tt torchdiffeq}\citep{torchdiffeq}. The generation results are measured by negative log-likelihood (NLL) in Table \ref{NLLS2}, which are computed from solving the likelihood ODE, parallel with the sampling ODE, by the adjoint method implemented by {\tt torchdiffeq}, and the generative visualization is shown in Figure \ref{Flood36}.

\begin{table}[htbp]
    \caption{Negative log-likelihood ($\downarrow$) results on four earth datasets. The best result is marked in bold. All numbers are averaged over 5 runs of experiments and each run contains 10 rounds of sampling. Each time of sampling generates samples of the same size as that of the primary datasets. The diffusion coefficient is fixed at $\epsilon=0.01$ and the likelihood step is adaptive for Dopri5 solver.}
    \vspace{5pt}
    \centering
    \begin{tabular}{lrrrr}
        \toprule
        Dataset  & Volcano & Earthquake & Flood & Fire  \\
        Size     & 827     & 6120       & 4875  & 12810 \\
        \midrule
        RCNF \citep{RCNF} & $-6.05\pm0.61$ & $0.14\pm0.23$ & $1.11\pm0.19$ & $-0.80\pm0.54$ \\
        Moser Flow \citep{MF} & $-4.21\pm0.17$ & $-0.16\pm0.06$ & $0.57\pm0.10$ & $-1.28\pm0.05$ \\
        CNFM \citep{MCNF} & $-2.38\pm0.17$ & $-0.38\pm0.01$ & $0.25\pm0.02$ & $-1.40\pm0.02$ \\
        RFM \citep{FMG} & $-7.93\pm1.67$ & $-0.28\pm0.08$ & $0.42\pm0.05$ & $-1.86\pm0.11$ \\
        M-FFF \citep{MFFF}& $-2.25\pm0.02$ & $-0.23\pm0.01$ & $0.51\pm0.01$& $-1.19\pm0.03$ \\
        \midrule
        StereoSGM \citep{RSGM} & $-3.80\pm0.27$ & $-0.19\pm0.05$ & $0.59\pm0.07$ & $-1.28\pm0.12$ \\
        RSGM \citep{RSGM} & $-4.92\pm0.25$ & $-0.19\pm0.07$ & $0.45\pm0.17$ & $-1.33\pm0.06$ \\
        RDM \citep{RDM} & $-6.61\pm0.97$ & $-0.40\pm0.05$ & $0.43\pm0.07$ & $-1.38\pm0.05$ \\
        RSGM-improved \citep{RSGMI} & $-4.69\pm0.29$ & $-0.27\pm0.05$ & $0.44\pm0.03$ & $-1.51\pm0.13$\\
        LogBM \citep{RDB} & $-9.52\pm0.87$ & $-0.30\pm0.06$ & $0.42\pm0.08$ & $\mathbf{-2.47\pm0.11}$ \\
        \midrule
        RNGI(Ours) & $\mathbf{-13.03\pm0.64}$ & $\mathbf{-0.79\pm0.08}$ & $\mathbf{0.10\pm0.07}$ & $\mathbf{-2.31\pm0.05}$ \\
        \bottomrule
    \end{tabular}
    \label{NLLS2}
\end{table}

Under different settings of diffusion intensity coefficient $\epsilon$, the proposed E-SDE algorithm keeps more fine-grained features compared to ODE sampler and outperforms conventional spherical GRW scheme in stochastic generative quality without sacrificing on computing efficiency. As shown in Table \ref{disS2}, the larger choice of $\epsilon$ perturbs the sampling results from the GRW scheme more seriously than those from the E-SDE scheme, which illustrates the improvement of sampling quality by the global construction of the diffusion part of E-SDE.

The weakness of local simulation used in GRW scheme can also be obvious when there are more discretzation steps. Table \ref{step} shows the results of the ablation study of $N$, where the sampling precision of E-SDE scheme benefits from the finer stepsize, but the performance of GRW scheme conversely decreases, which suffers from the inaccurate Gaussian stochastic step magnified by too much exponetial projections.

\begin{table}[htbp]
    \caption{Wasserstein-2 distance ($\downarrow$) between the ground truth distribution and the generated distribution. The sampling steps are fixed at $N=100$.}
    \vspace{5pt}
    \centering
    \begin{tabular}[width=\textwidth]{llrrrr}
    \toprule
    \multirow{2}{*}[-3pt]{{\small Sampler}} & \multirow{2}{*}[-3pt]{$\epsilon$} & \multicolumn{4}{c}{Dataset}\\
                  \cmidrule{3-6}
                  &    & \small{Volcano} & \small{Earthquake} & \small{Flood} & \small{Fire}  \\
                  \midrule
    ODE           &0     & 0.3154 & 0.2088 & 0.2141 & 0.1729 \\
                  \midrule
                  &0.1   & 0.8180 & 0.7432 & 0.7588 & 1.0596 \\
    GRW           &0.01  & 0.5467 & 0.5006 & 0.5321 & 0.6466 \\
                  &0.001 & 0.3674 & 0.2952 & 0.3178 & 0.3546 \\
                  \midrule
                  &0.1   & 0.4299 & 0.3460 & 0.3122 & 0.6883 \\
    E-SDE         &0.01  & 0.3326 & 0.2084 & 0.2095 & 0.2920 \\
                  &0.001 & 0.3112 & 0.2035 & 0.2147 & 0.1583 \\
                  \bottomrule
    \end{tabular}
    \label{disS2}
\end{table}

\begin{table}[t]
    \caption{Wasserstein-2 distance ($\downarrow$) between the ground truth distribution and the generated distribution. Generated samples have the same size as that of the primary datasets. The diffusion coefficient of GRW and E-SDE is fixed at $\epsilon=0.01$.}
    \vspace{5pt}
    \centering
    \begin{tabular}{llccccr}
    \toprule
    \multirow{2}{*}[-3pt]{Sampler} & \multirow{2}{*}[-3pt]{Steps} & \multicolumn{4}{c}{Dataset} \\
                  \cmidrule{3-6}
                  &    & \small{Volcano} & \small{Earthquake} & \small{Flood} & \small{Fire}  \\
                  \midrule
                  &10     & 0.3680 & 0.2578 & 0.2572 & 0.2637 \\
    ODE           &100    & 0.3154 & 0.2088 & 0.2141 & 0.1729 \\
                  &1000   & 0.3074 & 0.1975 & 0.2118 & 0.1742 \\
                  \midrule
                  &10     & 0.3923 & 0.3055 & 0.3291 & 0.4569 \\
    GRW           &100    & 0.5467 & 0.5006 & 0.5321 & 0.6466 \\
                  &1000   & 0.8375 & 0.7486 & 0.7679 & 0.9543 \\
                  \midrule
                  &10     & 0.4493 & 0.2528 & 0.2453 & 0.3389 \\
    E-SDE         &100    & 0.3326 & 0.2084 & 0.2095 & 0.2920 \\
                  &1000   & 0.2938 & 0.2114 & 0.2044 & 0.2947 \\
                  \bottomrule
    \end{tabular}
    \label{step}
\end{table}

\subsection{Density generation on \textbf{SO}(3)}

In this experiment, we bridge the uniform distribution and various synthetic probability distributions $\mu_i(i=1,\dots, 7)$ on \textbf{SO}(3). There are four target densities named as peak, cube, cone, line, which are designed from the symmetry of exact geometry objects in $\mathbb{R}^3$, and three multi-modal target densities $\mathbb{W}^{16}, \mathbb{W}^{32}, \mathbb{W}^{64}$ which distribute as 'Wrapped Gaussian' on \textbf{SO}(3).

%We train three RNGI models with different settings. RNGI-D model does not use any embedding and is directly parameterized on \textbf{SO}(3) while both of RNGI-EM and RNGI-ES models rely on the 'truncation-orthogonalization' embedding of target distributions $\mu_i$ into $\mathbb{R}^6$ as $\bar{\mu}_i= \mu_i\circ\mathcal{P}^{-1}$. The difference between RNGI-EM and RNGI-ES model is that the initial distribution of RNGI-EM is embedded $\bar{U}[\mathbf{SO}(3)] = {U}[\mathbf{SO}(3)] \circ \mathcal{P}^{-1}$ and that of RNGI-ES is directly $U(\mathbb{S}^5)$.
We train three RNGI models with different representation methods of the velocity and score field. As the element on \textbf{SO}(3) is represented by a $3\times 3$ matrix, the RNGI-D model learns the maps between the $3\times 3$ matrices, and then project the output matrix onto \textbf{SO}(3). The RNGI-EM and RNGI-ES models rely on the 'truncation-orthogonalization' embedding \citep{R62019} of target distributions $\mu_i$ into $\mathbb{R}^6$ as $\bar{\mu}_i= \mu_i\circ\mathcal{P}^{-1}$. The difference between RNGI-EM model and RNGI-ES model is that the initial distribution of RNGI-EM is embedded as $\bar{U}[\mathbf{SO}(3)] = {U}[\mathbf{SO}(3)] \circ \mathcal{P}^{-1}$ and that of RNGI-ES is directly $U(\mathbb{S}^5)$.

\begin{table}[htbp]
    \centering
    \caption{Log-likelihood ($\uparrow$) results(1) in density generation experiments on $\mathbf{SO}(3)$. The log-likelihood of initial samples are converted to 0 uniformly. All numbers are averaged over 5 runs of experiments and each run contains 10 rounds of sampling. Each time of sampling generates samples of the same size as that of the primary datasets. All of ODEs are numerically solved by dopri5 solver.}
    \vspace{5pt}
    \begin{tabular}{lrrrr}
        \toprule
        Dataset   & peak   & cube & cone & line   \\
        Size      & 600000& 600000& 600000& 600000\\
        \midrule
        RCNF\citep{RCNF}  & 13.47 & 1.02 & 8.82 & -0.0026  \\
        RELIE\citep{ReLie}  & 0.00 & 3.27 & 5.32 & -6.97  \\
        IPDF\citep{IPDF} & 7.30 & 4.33 & 4.75 & 1.12 \\
        Mixture-MF\citep{Fisher} & 10.52 & 4.52 & 8.36 & 0.77 \\
        Moser Flow\citep{MF} & 11.15 & 4.42 & 8.22 & 1.38  \\
        $\mathbf{SO}(3)$-NF(Mobius)\citep{CVPR} &13.93 &4.81 &8.98 &1.38 \\
        $\mathbf{SO}(3)$-NF(Affine)\citep{CVPR} &13.50 &0.00 &8.84 &0.00 \\
        \midrule
        RNGI-D & 14.52 & 1.84 & 6.05 & 0.00 \\
        RNGI-EM & 20.46 & 5.06 & 12.08 & 2.87\\
        RNGI-ES & \textbf{26.27} & \textbf{14.16} & \textbf{19.76} & \textbf{10.15} \\
        \bottomrule
    \end{tabular}
    \label{tab:NLLSO31}
\end{table}

While training RNGI-D model, we parameterize the velocity and score field with two fully-connected neural networks $v^{\theta}(t,x)$ and $s^{\eta}(t,x)$ whose depth is 6 and width is 512. The spatial input of $v^{\theta}$ are the 9-d flatten vector of the $3\times 3$ matrix for stronger expressivity, and that for $s^{\eta}$ are 3-d axis-angle vector of 3-d rotation matrix for more efficient divergence computation. Both of the dimensionality of the output of $v^{\theta}(t,x)$ and $s^{\eta}(t,x)$ are as same as the input. Then we optimize the networks with linear-scheduled AdamW and train 50000 iterations. In the sampling stage, we directly apply Dopri5 solver in {\tt torchdiffeq} package to solve matrix-valued ODE \ref{ODE}, and implement the geometric Euler-Maruyama solver for matrix-valued SDE in \texttt{pytorch}.

For RNGI-EM and RNGI-ES models, the parameterization of velocity and score field is established on $\mathbb{S}^5$. Therefore the two neural vector field, similarly parameterized by MLPs of size $6\times 512$, take 6-d representation both as their input and output, whose expressivity and computational effiency come from the flat geometry of $\mathbb{R}^6$. The optimizers and samplers are the same as RNGI-D, besides that the solver for E-SDE is Heun's. Both of the detailed parameter settings used for RNGI-D, RNGI-EM and RNGI-ES models are listed in Table \ref{paraso3} in the appendix \ref{subappendix:para}.

The generative visualization are shown in the appendix \ref{subappendix:vis}, where the visualization follows \citep{IPDF,CVPR}. Particularly, when sampling via ODE scheme, we use the adjoint method in \citep{RCNF} to evaluate the log-likelihood as quality measure. The results are reported in shown in Table \ref{tab:NLLSO31} and Table \ref{tab:NLLSO32} with baselines taken from \citep{CVPR} and \citep{MFFF}, which shows that both of RNGI-EM and RNGI-ES models reach higher log-likelihood than baseline methods, and RNGI-ES model reports the best performance.

\begin{table}[t]
    \caption{Log-likelihood ($\uparrow$) results(2) in density generation experiments on $\mathbf{SO}(3)$. The log-likelihood of initial samples are converted to 0 uniformly. All numbers are averaged over 5 runs of experiments and each run contains 10 rounds of sampling, and $\pm$ represents the standard deviation of the interval with a $95\%$ confidence level. Each time of sampling generates samples of the same size as that of the primary datasets. All of ODEs are numerically solved by dopri5 solver.}
    \vspace{5pt}
    \label{tab:NLLSO32}
    \centering
    \begin{tabular}{lrrr}
        \toprule
        Dataset  & $\mathbb{W}^{16}$ & $\mathbb{W}^{32}$ & $\mathbb{W}^{64}$ \\
        Size     & 600000 & 600000 & 600000 \\
        \midrule
        Moser Flow\citep{MF}  & $5.21\pm0.03$ & $4.53\pm0.03$ & $3.87\pm0.02$\\
        $\mathbf{SO}(3)$-NF\citep{CVPR} & $5.17\pm0.01$ & $4.48\pm0.004$ & $3.75\pm0.01$\\
        M-FFF\citep{MFFF} & $5.23\pm0.02$ & $4.57\pm0.02$ & $3.91\pm0.02$ \\
        \midrule
        StereoSGM\citep{RSGM} & $5.23\pm0.04$ & $4.52\pm0.03$ & $3.78\pm0.04$\\
        RSGM\citep{RSGM} & $5.25\pm0.03$ & $4.56\pm0.03$ & $3.87\pm0.02$ \\
        \midrule
        RNGI-D  & $3.04\pm0.03$  & $2.56\pm0.02$  & $2.01\pm0.03$ \\
        RNGI-EM & $7.76\pm0.04$  & $5.43\pm0.03$  & $3.98\pm0.01$ \\
        RNGI-ES & $\mathbf{14.94\pm0.05}$ & $\mathbf{13.89\pm0.04}$ & $\mathbf{12.64\pm0.05}$ \\
        \bottomrule
    \end{tabular}
    \end{table}

The RNGI-EM/ES models can not only generate new samples with higher accuracy, but also produce less computational cost in both training and sampling stage. One single training iteration of RNGI-D model takes 2.5s - 3s on a NVIDIA RTX 3090 GPU but that of RNGI-EM/ES model only takes 0.025s - 0.4s, because the flatten representation used by the velocity network and the axis-angle representation used by the score network of RNGI-D model make the loss surface stiff and the gradient computation harder, therefore slow down the back propagation. In the sampling stage, to generate $600000\times 20\% = 120000$ new samples, RNGI-E model takes 9.5s and 4.3s using E-SDE and ODE schemes respectively, and the time cost for the GRW and ODE schemes of RNGI-D model are 12.6s and 5.6s. Note that RNGI-E and RNGI-D models take 6-d vector and $3\times 3$ matrix as their data format separately, thus the former consumes a third less memory usage compared to the latter.

\subsection{Interpolation between mixture densities}

Another advantage of RNGI models is that it does not restrict the sampling process to start from the uniform distribution or any projected distribution from a single tangent space like wrapped Gaussian distribution \citep{wrapped} as traditional diffusion models. To demonstrate the flexibility on the initial distribution, we also use the RNGI models to interpolate between the two multi-modal densities with different modals on $\mathbb{S}^2$ and \textbf{SO}(3). Intuitively, the models are trained to push different density centers to merge or pull them to split, which is a challenging task due to the non-flat geometry of $\mathbb{S}^2$ and \textbf{SO}(3).

On $\mathbb{S}^2$, we train RNGI model to connect two $K$-center VMF mixture distributions $\mathbb{V}^K$ \citep{wrapped}. The setting of two endpoint distributions are $\mathbb{V}^{32}\leftrightarrow \mathbb{V}^{8}$, $\mathbb{V}^{64}\leftrightarrow \mathbb{V}^{8}$, $\mathbb{V}^{64}\leftrightarrow \mathbb{V}^{16}$ and $\mathbb{V}^{32}_{\{\mu_i\}}\to\mathbb{V}^{32}_{\{\nu_i\}}$ respectively, where $\leftrightarrow$ means the generation is conducted for both directions. The $K$ centers of $\mathbb{V}^K$ are random points on $\mathbb{S}^2$ that are drawn from $U(\mathbb{S}^2)$ and all of the concentration parameters $\kappa$ are uniformly set to 256. As for $\mathbf{SO}(3)$, we train RNGI-D and RNGI-E model for the interpolation between wrapped $K$-mixture Gaussian distributions $\mathbb{W}^K$. The setting of $K$ for $\mathbb{W}^K$ is the same as that of $\mathbb{V}^K$, and the position of centers and scaling parameters are $p_i\sim U[\mathbf{SO}(3)]$ and $\sigma_i^2=0.01$.

We use MLPs of size $6\times 1024$, who are also optimized by linear-scheduled AdamW with 50000 iterations, to paramerize the two vector fields in both $\mathbb{S}^2$ and \textbf{SO}(3) experiments. All of the parameter settings of network training are same and shown in Table \ref{paraso3} in Appendix \ref{subappendix:para}. And we calculate the $\mathrm{K-L}$ divergence between ground truth distribution and generated distribution as the metric of precision. 

The bi-directional numerical results and generative visualization are collected in Appendix \ref{subappendix:dis} and \ref{subappendix:vis}. As shown in them, both of RNGI model on $\mathbb{S}^2$ and RNGI-D/E models on \textbf{SO}(3) can connect and convert two complex densities on their living manifolds. On $\mathbb{S}^2$, the E-SDE scheme generates the most realistic two endpoint VMF mixture densities, where ODE scheme distorts the centers into 'ribbons' and GRW scheme generates many points out of distributions. And on \textbf{SO}(3), while the training of RNGI-D model is similarly much slower than RNGI-E model, both models capture the main feature of the two multl-modal endpoint densities. The generative style of the two models are slightly different that RNGI-D are more serrated and RNGI-E are more diffusive, and RNGI-E model and its E-SDE scheme suffer from much less computational burden as their training is not as computationally-intensive as RNGI-D model and its GRW scheme.

\section{Conclusion}
In this work, we propose \textit{Riemannian Neural Geodesic Interpolant}, a generative model capable of bridging two arbitrary probability distributions on Riemannian manifolds. Based on the RNGI, we design learning and sampling algorithms for density generation on Riemannian manifold $\mathbb{S}^2$ and \textbf{SO}(3). Our model shows significant flexibility in bridging complex distributions and exhibits observable improvement in sampling quality with the E-SDE sampler. Furthermore, the RNGI model still has the potential to be practically deployed in broader generation tasks defined on various Riemannian manifolds. For instance, we will extend the RNGI from point generation to point cloud generation on general submanifolds in $\mathbb{R}^d$ in the future.

\section*{Acknowledgments}
We express thanks to Xiangdong Li, Fuzhou Gong and Xicheng Zhang for the mathematical discussions, and Hongsheng Qi for his support on the computational resources.

\newpage
\bibliography{main}
\bibliographystyle{tmlr}

\newpage
\appendix

\section{Stochastic Differential Geometry}\label{appendix:SDG}
In this section, we list some fundamental concepts and notations in stochastic differential geometry.

We begin with some preliminaries of smooth manifold. And see \cite{Lee13} for a more detailed and comprehensive account. A smooth d-dimensional manifold is a topological space $\mathcal{M}$ and a family of pairs (called charts) $\{(U_i, \varphi_i)\}$, where the $\{U_i\}$ are open cover of $\mathcal{M}$ and $\varphi_i$ is a homeomorphism from $U_i$ to an open subset of $\mathbb{R}^d$. The charts are required to satisfy a compatibility condition: if $U_i\cap U_j=U$, then $\varphi_i\circ\varphi_j^{-1}|_U$ is a smooth map from $\varphi_j(U)$ to $\varphi_j(U)$. The set of smooth functions on $\mathcal{M}$ is denoted $C^\infty(\mathcal{M})$ whose element has type $f: \mathcal{M}\rightarrow \mathbb{R}$ such that for any chart $(U,\varphi)$ the map $f\circ\varphi^{-1}$ is smooth. 

\begin{definition}
For a given point $x\in\mathcal{M}$, a derivation at $x$ is a linear operator $D: C^\infty(\mathcal{M})\rightarrow \mathbb{R}$ such that for all $f,g\in C^\infty(\mathcal{M})$, 
$$D(fg)=f(x)D(g)+g(x)D(f).$$
The  derivation at $x$ is also called the tangent vector  at $x$.   The set of all  tangent vectors is a d-dimensional real vector space called the tangent space $T_x\mathcal{M}$. The tangent bundle is denoted by $T\mathcal{M}$,  consists of the tangent space $T_x\mathcal{M}$ at all points $x$ in $\mathcal{M}$:
\begin{align*}
 T\mathcal{M}&=\bigsqcup_{x\in\mathcal{M}} T_x\mathcal{M}=\{(x, v_x)| x\in\mathcal{M}, v_x\in T_x\mathcal{M}\} .
\end{align*}
 
\end{definition}

Every smooth manifold can be embedded in  $\mathbb{R}^m$ with $m>d$ for some suitably chosen $m$. We can view $T_x\mathcal{M}$ as a linear subspace of $T_x\mathbb{R}^m$, thus a tangent vector can be written as: 
$$D=\sum^d_{j=1}d_j \frac{\partial}{\partial x_j}.$$

A vector field $X$ is a continuous map such that for each point $x$ on the manifold,  $X(x)\in T_x\mathcal{M}$.  Such a vector field can also map any smooth function $f$ to a function, via the assignment
$x\in \mathcal{M}\rightarrow X(x)(f)\in \mathbb{R}$. If $X(\cdot)(f)$ is smooth we say that the vector
field is smooth. The space of smooth vector fields on $\mathcal{M}$ is denoted  by $\mathcal{X}(\mathcal{M})$.

Every smooth manifold $\mathcal{M}$ can be equipped with a Riemannian metric $g$. $g$ is a metric tensor field such that for each $x\in\mathcal{M}$, $g$ defines an inner product  $g(x): T_x\mathcal{M}\times T_x\mathcal{M}\rightarrow \mathbb{R}$ on the tangent space.  For any $X, Y \in \mathcal{X}(\mathcal{M})$, we denote $\langle X, Y \rangle_g = g(X, Y)$ with $g(X, Y)(x)=g_x(X(x), Y(x))$. Note that in local coordinates we can define 
$G=\{g_{ij}\}=\{g(X_i, X_j)\}$ where $\{X_{i}\}_{1\leqslant i\leqslant d}$ is a basis of the
tangent space. This allows us to write the metric using the coordinates 
$$\langle X, Y \rangle_g =\sum_{i,j}x_i y_j g_{ij}.$$

The Riemannian metric allows us to define a measure over measurable subsets of the manifold. For a single chart $(U,\varphi)$ and any smooth function $f$ supported in U, consider the following positive linear functional: 
$$f\longmapsto \int_{\varphi(U)}(f\sqrt{|\mathbf{det}G|})\circ \varphi^{-1}dx=\int_U fdM_{U},$$
where $M_U$ is  a unique Borel measure given by Riesz representation theorem. Then  a partition of
unity method allows us to extend $M_U$  to be defined over the entire $\mathcal{M}$, which leads to the Riemannian measure $M$. A probability density function  $p$ over $\mathcal{M}$ can be thought of as a non-negative  function satisfying $\int_{\mathcal{M}}pdM=1$.

A crucial structure closely related to the Riemannian metric is the gradient operator on $\mathcal{M}$. For any $f\in C^\infty(\mathcal{M})$ and $Y\in \mathcal{X}(\mathcal{M})$, the gradient operator $\nabla$ is defined via
$\langle \nabla f, Y\rangle_g=Y(f)$. The divergence operator $\mathbf{div}$ (or $\nabla\cdot$) can be obtained via the following Stokes formula: for any $f\in C^\infty(\mathcal{M})$ and any $X\in \mathcal{X}(\mathcal{M})$, $\int_{\mathcal{M}} \mathbf{div}X fdM =-\int_M X(f)dM$. The Laplace-Beltrami operator $\Delta_{\mathcal{M}}$ is given by $\Delta_{\mathcal{M}}f:=\mathbf{div}(\nabla f)$.

We now turn to stochastic differential equations on $\mathcal{M}$.  Solutions of stochastic differential equations on manifolds should be sought in the space of manifold-valued semimartingales. A continuous $\mathcal{M}$-valued stochastic process $X_t$ is called a $\mathcal{M}$-valued semimartingale if for any $f\in C^\infty(\mathcal{M})$ we have that $f(X_t)$ is a real valued semimartingale.

A stochastic differential equation on $\mathcal{M}$ is defined by $l$ vector fields $V_1,\dots, V_l\in \mathcal{X}(\mathcal{M})$, an $\mathcal{R}^l$-valued  semimartingale $Z$, and an $\mathcal{M}$- valued random variable $X_0$, serving as the initial value of the solution. For more details we refer to \cite{PeiHsu}.

\begin{definition}
A $\mathcal{M}$-valued semimartingale $X_t$ is said to be the solution of  $SDE(V, Z; X_0)$ up to a
stopping time $\tau$ if for any $f\in C^\infty(\mathcal{M})$ and $t\in [0, \tau]$, 
$$f(X_t)=f(X_0)+\int^t_0\sum^l_{i=1}V_i(f)(X_s)\circ dZ_s^i.$$
\end{definition}

Using the Laplace–Beltrami operator, we can give the definition of the Brownian motion on $\mathcal{M}$ as a diffusion process. 

\begin{definition}
Let $B^{\mathcal{M}}_t$ be a $\mathcal{M}$-valued semimartingale. We say $B^{\mathcal{M}}_t$ is a Brownian motion on $\mathcal{M}$ if for any $f\in C^\infty(\mathcal{M})$, the following process is a real local martingale:
\begin{align*}
    f(B^{\mathcal{M}}_t)-f(B^{\mathcal{M}}_0)-\frac12\int^t_0 \Delta_{\mathcal{M}}f(B^{\mathcal{M}}_s)ds.
\end{align*}
\end{definition}

This definition considers Brownian motion as a diffusion process generated by the Laplace-Beltrami operator $\frac{\Delta_{\mathcal{M}}}{2}$, and is shown to be equivalent to several other descriptions, including the one stating that it is a semimartingale whose anti-development with the Levi-Civita connection is a Euclidean Brownian motion. Meanwhile, it is much more convenient to use equivalent characterizations such as embedding Euclidean brownian motion or Stroock's representation when considering realization.

\section{Theoretical guarantee of RNGI}\label{appendix:prf}

\paragraph{Proof of Theorem \ref{Thm}}  For a wide range of test function $\varphi(x)$ on $\mathcal{M}$, we have 
   $$\mathbb{E}\varphi(x_t) = \int_\mathcal{M} \varphi(x)\rho(t,x)\mathrm{d}M(x),$$
   $$\frac{\mathrm{d}}{\mathrm{d}t} \mathbb{E}\varphi(x_t)  = \int_\mathcal{M} \varphi(x) \partial_t\rho(t,x) \mathrm{d}M(x).$$

   At the same time, the time derivative of expectation can also be calculated as
   \begin{align*}
       \frac{\mathrm{d}}{\mathrm{d}t} \mathbb{E}\varphi(x_t) 
       &= \mathbb{E} [\frac{\mathrm{d}}{\mathrm{d}t} \varphi(x_t)] 
        = \mathbb{E} [\nabla \varphi \cdot \partial_t x_t ]\\
       &= \int_\mathcal{M} \nabla \varphi \cdot \mathbb{E}[\partial_t x_t|x_t = x] \rho(t,x) \mathrm{d}M(x)\\
       &= \int_\mathcal{M} \nabla \varphi \cdot v(t,x) \rho(t,x) \mathrm{d}M(x)\\
       &= \int_\mathcal{M} -\varphi \cdot \mathbf{div} v(t,x)\rho(t,x) \mathrm{d}M(x).
   \end{align*}
   Combine above two equation we have
   $$\int_\mathcal{M} \varphi \cdot \{\partial_t \rho(t,x) + \mathbf{div} [v(t,x)\rho(t,x)]\} \mathrm{d}M(x) = 0.$$

Having proved Theorem \ref{Thm}, we only need to verify the integrable condition on manifold we are concerned with to ensure our implement on them is theoretically solid. 

\paragraph{Hypersphere $\mathbb{S}^n$} 
We first state that Geodesic Interpolant is well-defined on almost the whole $\mathbb{S}^n$. Note that for $p\in\mathbb{S}^n$ and $v\in T_p\mathbb{S}^n$, when $|v|=\pi$, the corresponding point on $\mathbb{S}^n$ is $-p$ but the injective property does not hold anymore as the non-uniqueness of the shortest path between $p$ and $-p$. Therefore, only by considering the 'almost complete' alternative $\mathbb{S}^n-\{-p\}$, we can construct RNGI model. \par
Now back to our context of integrable condition. For the anti-projection, denoted as $\mathrm{Log}_{x_0}x_1$, of the final random variable $x_1$, which is a tangent vector living in $T_{x_0}\mathbb{S}^n$, 
We have given the expression of Geodesic Interpolant on $\mathbb{S}^n$ as
\begin{align*}
    I(t;x_0,x_1) &= \mathrm{Exp}_{x_0} (t \cdot \mathrm{Log}_{x_0}x_1) \\
                 &= \cos(t \cdot |\mathrm{Log}_{x_0}x_1|)x_0 + \sin(t \cdot |\mathrm{Log}_{x_0}x_1|)\frac{\mathrm{Log}_{x_0}x_1}{|\mathrm{Log}_{x_0}x_1|}.
\end{align*}
Then we can calculate the velocity field of the interpolating curve as:
\begin{equation*}
    \partial_tI(t;x_0,x_1) = \cos(t \cdot |\mathrm{Log}_{x_0}x_1|)\mathrm{Log}_{x_0}x_1
                           -|\mathrm{Log}_{x_0}x_1|\sin(t \cdot |\mathrm{Log}_{x_0}x_1|)x_0. 
\end{equation*}
Notice that $x_0$ and $\mathrm{Log}_{x_0}x_1$ are orthogonal vectors in $\mathbb{R}^{n+1}$, so
\begin{equation*}
    |\partial_t I(t;x_0,x_1)| = |\mathrm{Log}_{x_0} x_1| < 2\pi,
\end{equation*}
where '$<$' comes from the compactness of $\mathbb{S}^n$. And from the conditional expectation definition of $v(t,x)$ and full expectation formula we have
\begin{equation*}
    \int_{\mathcal{M}}|v(t,x)|\rho(t,x)\mathrm{d}M(x)\leq\mathbb{E}(|\partial_t I(t;x_0,x_1)|).
\end{equation*}
Combine above two inequalities we finally reach the integrable condition on $\mathbb{S}^n$:
\begin{equation*}
    \mathbb{E}(|\partial_t I(t;x_0,x_1)|) < \infty.
\end{equation*} 

\paragraph{(Matrix) Lie group} For theoretical completeness, we first introduce some basic knowledge about general Lie group. \par
On a general Lie group $G$ and its Lie algebra $\mathfrak{g}$, which can be seen as its tangent space at the identity element $e$,
the exponential map $\mathrm{Exp}$ is defined by the concept of single-parameter subgroup. \par
For any $X$ in Lie algebra $\mathfrak{g}$, let $\mathrm{exp}_X(t):\mathbb{R}\to G$ denotes the unique single-parameter subgroup whose derivative at $t=0$ is $X$ (the existance and uniqueness of such subgroup can be found in any textbook of Lie group theory), the exponential map is defined as
\begin{equation*}
    \mathrm{Exp}(X) = \mathrm{exp}_X(1),
\end{equation*}
and the logarithm map $\mathrm{Log}$ is locally defined as the inverse map of $\mathrm{Exp}$ without direct definition. And for those concerning Lie groups take matrices as their elements, the exponential and logarithm map have more explicit form
\begin{align*}
    &\mathrm{Exp}(X) = \sum_{k=0}^\infty \frac{1}{k!}X^k, \\
    &\mathrm{Log}(R) = \sum_{k=1}^\infty \frac{(-1)^{k+1}}{k} (R-I)^k,
\end{align*}
which are formally given by the exponential and power series of matrices and consistent with their name. Note that the symbol $\mathrm{Exp}$ and $\mathrm{Log}$ without the lower index always refer to the \textrm{Exp/Log} map at $e$, and then, carried by $\mathfrak{g}=T_eG$, be translated to any point $p$ by the group action induced by $p$.\par 
Particularly, the Rodrigues' rotation formula gives the connection between 3-D rotation vector and 3-D orthogonal matrix, which can be used to construct the bijective map between \textbf{SO}(3) and $\mathfrak{so}(3)$:
\begin{equation}\label{Exp}
    \mathrm{Exp}_e(\omega) = I + \sin\theta \cdot \omega + (1-\cos\theta) \cdot \omega^2,
\end{equation}
where $\theta = \sqrt{\omega_1^2+\omega_2^2+\omega_3^2}$ denotes the norm of $\omega$ (Frobenius for $\omega$ and Euclidean for $\hat{\omega}$). And inversely, the logarithm map from \textbf{SO(3)} to $\mathfrak{so}(3)$ at $e$ is given by:
\begin{equation}\label{Log}
    \mathrm{Log}_e
    \left[
    \begin{array}{ccc}
    R_{11} & R_{12} & R_{13} \\
    R_{21} & R_{22} & R_{23} \\
    R_{31} & R_{32} & R_{33}
    \end{array}
    \right]
    = \frac{\gamma}{2\sin\gamma}
    \left[
    \begin{array}{c}
    R_{32}-R_{23}  \\
    R_{13}-R_{31}  \\
    R_{21}-R_{12}
    \end{array}
    \right],
\end{equation}
where the rotation matrix $R\in\mathbf{SO}(3)$ is expressed by entries and rotation angle $\gamma$ is calculated by
\begin{equation*}
    \gamma = \arccos(\frac{\mathrm{tr}(R)-1}{2}).
\end{equation*}
Similarly to $\mathbb{S}^n$, we also state that how we construct RNGI on almost the whole \textbf{SO}(3). The orthogonality of \textbf{SO}(3) elements has a constraint on their trace that $\mathrm{tr}(R)\in[-1,3]$, so choose the main value of $\arccos$ then we get a mapped $\gamma\in[0,\pi]$. But for those matrices with $\mathrm{tr}=-1$, which represents a rotation with $\mathrm{angle}=\pi$, the bijection between rotation vectors and orthogonal matrices does not hold anymore, leading to {\tt nan} value of $\frac{\gamma}{2\sin\gamma}$. Now we denote the set $\{R\in\mathbf{SO}(3)|\mathrm{tr}(R)=-1\}$ and $\{\omega\in\mathfrak{so}(3)||\omega|<\pi\}$ as $R(-1)$ and $\omega(\pi)$, the bijective  $\mathrm{Exp}$ and $\mathrm{Log}$  can be constructed between
\begin{align*}
    &\mathrm{Exp}: \omega(\pi) \to [\mathbf{SO}(3)-R(-1)]\subset\mathbf{SO}(3),\\
    &\mathrm{Log}: [\mathbf{SO}(3)-R(-1)] \to\omega(\pi)\subset\mathfrak{so}(3).
\end{align*}
Every point $p\in$\textbf{SO}(3) gives a group-translation by group multiplication $p\cdot \mathbf{SO}(3)$, while the tangent space $T_p\textbf{SO}(3)$ at $p$ is concurrently given by the $p$-translation of Lie algebra. i.e.
\begin{equation*}
    p \cdot T_e\mathbf{SO}(3) = T_p\mathbf{SO}(3).
\end{equation*}
And for any tangent vector $v\in T_p\mathbf{SO}(3)$ and any other point $q\in\mathbf{SO}(3)$, their relative position of $p$ is given by $p^{-1}\cdot(p,v,q) = (e,p^{-1}v,p^{-1}q)$. Then the $\mathrm{Exp}$ and $\mathrm{Log}$ map at $p$ are translated from those at $e$:
\begin{align*}
    &\mathrm{Exp}_p(v) = p\cdot \mathrm{Exp}_e(p^{-1}v),\\
    &\mathrm{Log}_p(q) = p\cdot\mathrm{Log}_e(p^{-1}q),
\end{align*}
For the random variable $x_0$ and $x_1$, the geodesic interpolant $I(t;x_0,x_1)$ on \textbf{SO}(3) is expressed as
\begin{align*}
    I(t;x_0,x_1) &= \mathrm{Exp}_{x_0} (t\cdot \mathrm{Log}_{x_0}x_1) \\
                 &= x_0\cdot \mathrm{Exp}_e [t \cdot \mathrm{Log}_e(x_0^{-1}x_1)].
\end{align*}

We can easily verify that $\mathrm{Exp-Log}$ map is length-preservable. So as $\mathrm{Log}_e(x_0^{-1}x_1)\in\mathfrak{so}(3)$, the time-derivative of geodesic interpolant $I(t;x_0,x_1)$ is 
\begin{equation*}
    \partial_tI(t;x_0,x_1) = x_0\cdot \mathrm{Exp}_e [t \cdot \mathrm{Log}_e(x_0^{-1}x_1)]\cdot \mathrm{Log}_e(x_0^{-1}x_1).
\end{equation*}
The left-invariant translation is given by matrix multiplication $x_0\cdot \mathrm{Exp}_e [t \cdot \mathrm{Log}_e(x_0^{-1}x_1)]$, which is length-preserving, so we have
\begin{equation*}
    |\partial_tI(t;x_0,x_1)| = |\mathrm{Log}_e(x_0^{-1}x_1)|.
\end{equation*}
By the Rodrigues' rotation formula and main value constraint, $|\mathrm{Log}_e(x_0^{-1}x_1)| \in [0,\pi)$, then the expectation $\mathbb{E}(|\partial_t I(t;x_0,x_1)|)$ is also bounded, thus the integrable condition on \textbf{SO}(3) is verified.

\paragraph{Proof of proposition \ref{KL}}
    By the definition of KL-divergence, we can calculate the time-derivative of $ \mathrm{KL}(\rho(t)\|\hat{\rho}(t))$ as
   \begin{equation*}
       \frac{\mathrm{d}}{\mathrm{d}t} \mathrm{KL}(\rho(t)\|\hat{\rho}(t)) = 
       \frac{\mathrm{d}}{\mathrm{d}t} \int_\mathcal{M}\log \frac{\rho(t,x)}{\hat{\rho}(t,x)} \rho(t,x)\mathrm{d}M(x).
   \end{equation*}
   Commute the time-differential and integral symbol and apply the chain rule, and notice that
   \begin{equation*}
       \int_\mathcal{M}\partial_t \rho(t,x)\mathrm{d}M(x) = \partial_t \int_\mathcal{M} \rho(t,x)\mathrm{d}M(x) = \partial_t 1 = 0,
   \end{equation*}
   we have
   \begin{align*}
       \frac{\mathrm{d}}{\mathrm{d}t} \mathrm{KL}(\rho(t)\|\hat{\rho}(t)) 
       = &\int_\mathcal{M} \log \frac{\rho(t,x)}{\hat{\rho}(t,x)}\partial_t\rho(t,x)\mathrm{d}M(x)\\
       &- \int_\mathcal{M}  \frac{\rho(t,x)}{\hat{\rho}(t,x)}\partial_t\hat{\rho}(t,x)\mathrm{d}M(x).
   \end{align*}
   Substitute the two transport equations into above expression, and use the integration by parts formula, we have
   \begin{align*}
       &\frac{\mathrm{d}}{\mathrm{d}t} \mathrm{KL}(\rho(t)\|\hat{\rho}(t)) \\
       = &  \int_\mathcal{M}\langle \nabla\log \rho(t,x)-\nabla\log \hat{\rho}(t,x),v(t,x)\rangle_g\rho(t,x)\mathrm{d}M(x)\\
       & - \int_\mathcal{M}\langle\nabla\frac{\rho(t,x)}{\hat{\rho}(t,x)},\hat{v}(t,x)\rangle_g\hat{\rho}(t,x)\mathrm{d}M(x).
   \end{align*}
   Apply the chain rule of Riemannian gradient to the second term, we have
   \begin{align*}
       &\int_\mathcal{M}\langle\nabla \frac{\rho(t,x)}{\hat{\rho}(t,x)},\hat{v}(t,x)\rangle_g\hat{\rho}(t,x)\mathrm{d}M(x) \\
       = &\int_\mathcal{M}\langle\nabla \log \rho(t,x)-\nabla\log \hat{\rho}(t,x),\hat{v}(t,x)\rangle_g\rho(t,x)\mathrm{d}M(x).
   \end{align*}
   Merge above two equations and base on the bilinear property of Riemannian metric $g$, we finally get the exact expression of the time derivative
   \begin{align*}
       &\frac{\mathrm{d}}{\mathrm{d}t} \mathrm{KL}(\rho(t)\|\hat{\rho}(t)) \\
       =&\int_\mathcal{M} \langle s(t,x)-\hat{s}(t,x),v(t,x)-\hat{v}(t,x)\rangle_g \rho(t,x)
        \mathrm{d}M(x).
   \end{align*}
   
   Integration from $t=0$ to $t=1$,  by the fact that $\rho(0)$ and $\hat{\rho}(0)$ denote the same prior distribution so the initial KL-divergence $\mathrm{KL}(\rho(x,0)\|\hat{\rho}(x,0)) = 0$, the proof is completed.

Furthermore, we can model the discrepancy of the interpolating process and generating process in the sense of Wasserstein metric.

\paragraph{Proof of proposition \ref{WP}}
     For any $f\in C^\infty(\mathcal{M})$, 
   \begin{align*}
     |f(X_t)-f(\hat{X}_t)|\leq &\int^t_0|v(f)(X_s, s)-\hat{v}(f)(\hat{X}_s, s)|ds\\
     \leq &\int^t_0(|v(f)(X_s, s)-{v}(f)(\hat{X}_s, s)|+|v(f)(\hat{X}_s, s)-\hat{v}(f)(\hat{X}_s, s)|)ds\\
     \leq &\int^t_0(L|f(X_s)-f(\hat{X}_s)|+\sup_{s\in[0,1]}\sup_{x\in\mathcal{M}}|v(f)(x, s)-\hat{v}(f)(x, s)|)ds.
   \end{align*}    
 Gronwall’s lemma gives 
 $$ |f(X_t)-f(\hat{X}_t)|\leq e^{Lt} \sup_{s\in[0,1]}\sup_{x\in\mathcal{M}}|v(f)(x, s)-\hat{v}(f)(x, s)|.$$
 Hence let $d_g$ be the metric induced by $g$, we have
 $$d_g(X_t,\hat{X}_t)\leq e^{Lt} \sup_{s\in [0,t]}\sup_{x\in\mathcal{M}}|v(x,s) - \hat{v}(x,s)|_g.$$
 Then we have
 \begin{align*}
 W_p(\mu_t, \hat{\mu}_t) \leq &[\mathbb{E}d_g(X_t,\hat{X}_t)^p]^{\frac{1}{p}}\leq e^{Lt}
      \sup_{s\in [0,t]}\sup_{x\in\mathcal{M}}|v(x,s) - \hat{v}(x,s)|_g.
 \end{align*}

\section{Experiment Details}\label{appendix:exp}
\begin{figure}[htbp]
    \centering
    \includegraphics[width=\textwidth]{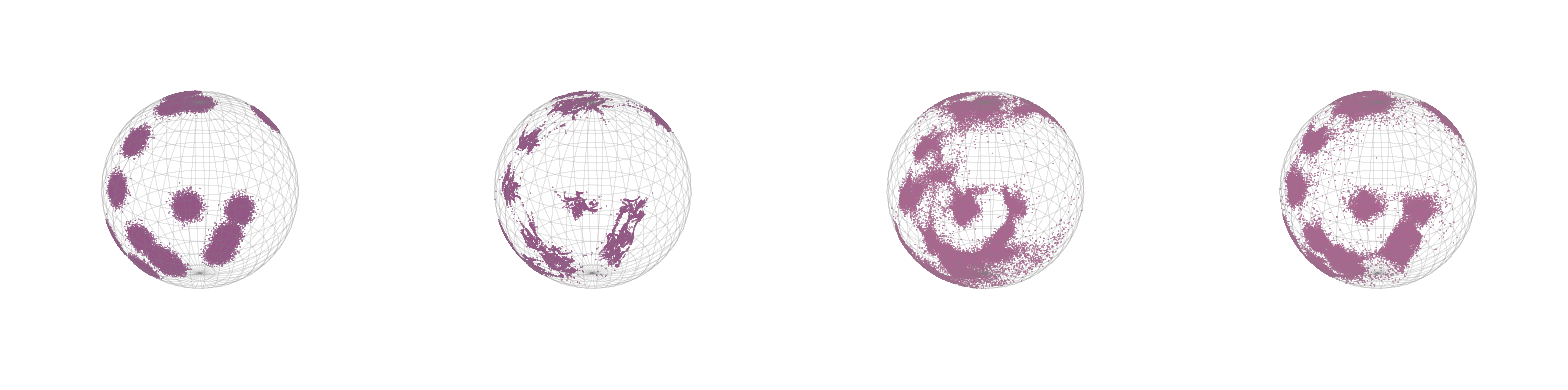}
    \caption{The generative performance of RNGI model on $\mathbb{S}^2$ in the distribution connection experiment $\mathbb{V}^{64}\to\mathbb{V}^{16}$. Both of the observation angle and the order of subfigure are the same as Figure \ref{Flood36}. }
    \label{fig:6416S2}
\end{figure}

\begin{figure}[htbp]
    \centering
    \includegraphics[width=\textwidth]{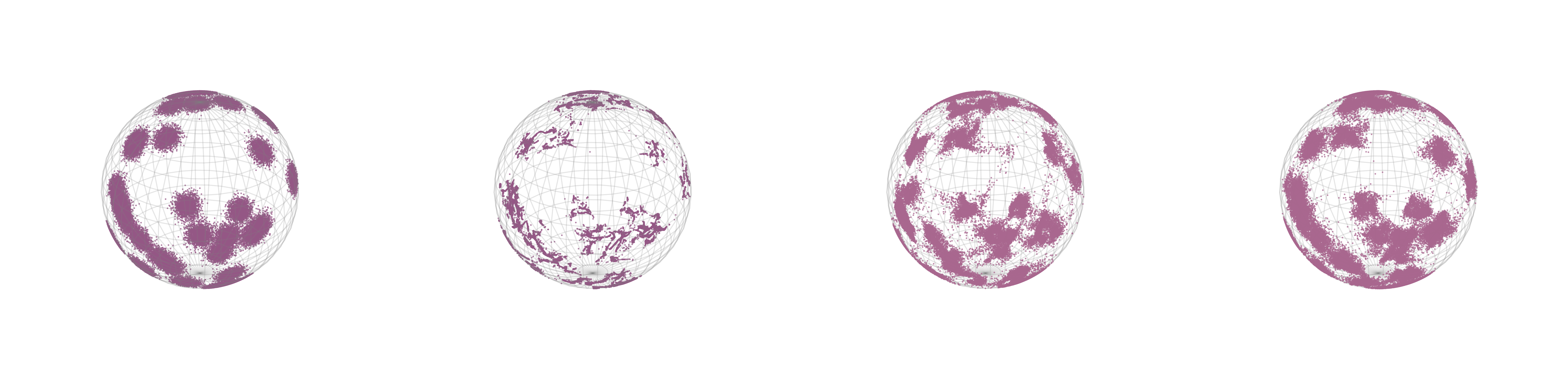}
    \caption{The generative performance of RNGI model on $\mathbb{S}^2$ in the distribution connection experiment $\mathbb{V}^{32}_{\{\mu_i\}}\to\mathbb{V}^{32}_{\{\nu_i\}}$. Both of the observation angle and the order of subfigure are the same as Figure \ref{Flood36}.}
    \label{fig:3232S2}
    \end{figure}

\begin{figure}[htbp]
    \centering
    \includegraphics[width=\textwidth]{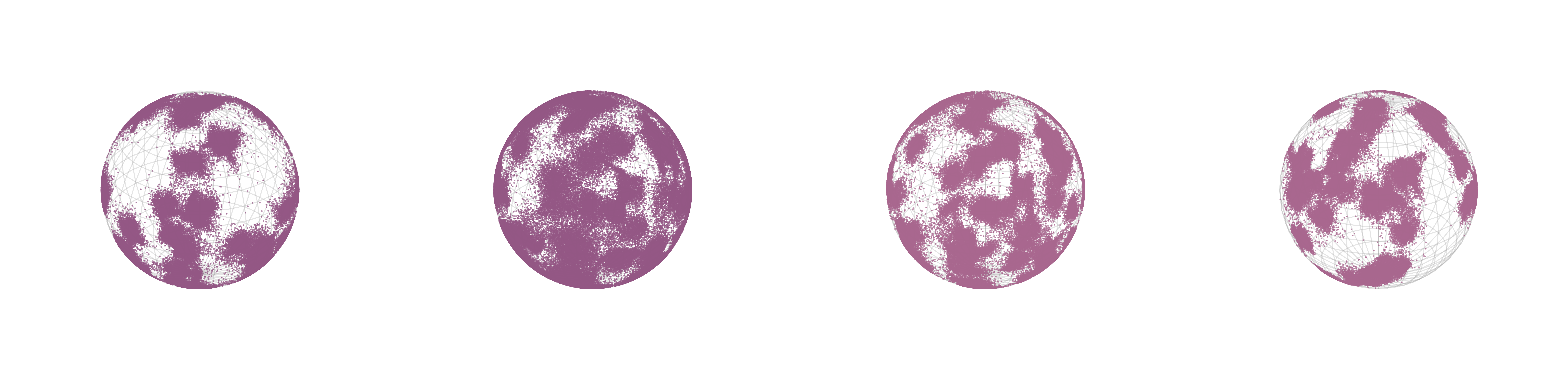}
    \caption{The backward generative performance by the E-SDE scheme of RNGI model on $\mathbb{S}^2$ in the four distribution connection experiments. From left to right are $\mathbb{V}^{8}\to\mathbb{V}^{32}$, $\mathbb{V}^{8}\to\mathbb{V}^{64}$, $\mathbb{V}^{16}\to\mathbb{V}^{64}$ and $\mathbb{V}^{32}_{\{\nu_i\}}\to\mathbb{V}^{32}_{\{\mu_i\}}$}
    \label{fig:VMFESDE0}
    \end{figure}

Below we list the hyperparameters used to set RNGI model in various experiments in Table \ref{paras2} and Table \ref{paraso3}, the generative performance metrics from Table \ref{tab:cwoS2} to Table \ref{tab:cdoEs}, and the generative visualization from Figure \ref{fig:6416S2} to Figure \ref{fig:d1o0}. We promise that all the experiments are reproducible and will release the code after the acceptance of the paper.

\subsection{Parameter setting}\label{subappendix:para}
Below the tables are the hyperparameters used to set the RNGI model in various experiments. Learning rate, weight decay, step and gamma, namely optimizing 
hyperparameters, are pre-searched by Optuna\citep{Optuna2019} and then uniformly decided. And the searching interval for the four optimizing 
hyperparameters are $[0.0001,0.01]$, $[0.001,0.01]$, $\{2000,2500,\dots,4500,5000\}$ and $\{0.4,0.45,\dots,0.85,0.9\}$ respectively.

\begin{table}[htbp]
    \centering
    \caption{Hyperparameters for the training of RNGI model on geographical datasets on $\mathbb{S}^2$. $a/b$ means $a$ for velocity training and $b$ for score training respectively and a single $a$ means $a$ for both velocity and score training.}
    \vspace{10pt}
    \begin{tabular}{l|cccc}
        \toprule
        Parameters  & Volcano & Earthquake & Flood & Fire \\
        \midrule
        Layers &5 & 5 & 5 & 7 \\ 
        Hidden units &512&512&512&512\\
        Activation &relu&relu&relu&tanh\\
        \midrule
        Training steps&100000&100000&100000&150000\\
        Batch size&512&512&512&512\\
        Optimizer&RiemannianAdam&RiemannianAdam&RiemannianAdam& AdamW\\
        Learning rate &0.003/0.0003&0.003/0.0003&0.003/0.0003&0.003/0.001\\
        Weight decay& - & - & - &0.004/0.01\\
        Scheduler&StepLR &StepLR &StepLR &StepLR \\
        Step&2500/1500&3000/1500&2500&4000/2500\\
        Gamma&0.7/0.4&0.8/0.4&0.7&0.8/0.7\\
        \bottomrule
    \end{tabular}
    \label{paras2}
\end{table}

\begin{table}[htbp]
    \centering
    \caption{Hyperparameters for the density generation experiment on \textbf{SO}(3) and modals interpolation experiments on both $\mathbb{S}^2$ and \textbf{SO}(3).}
    \vspace{10pt}
    \begin{tabular}{l|c|c}
        \toprule
        Parameters& Density generation & Modals interpolation \\
        \midrule
        Layers& 6 & 6 \\ 
        Hidden units& 512 & 1024\\
        Activation& relu & relu\\
        \midrule
        Training steps& 50000& 50000\\
        Batch size & 512 & 1024 \\
        Optimizer& AdamW & AdamW \\
        Learning rate& 0.0005 & 0.0005\\
        Weight decay& 0.01 & 0.01\\
        Scheduler& StepLR & StepLR\\
        Step& 2500 & 2500\\
        Gamma& 0.7 & 0.7\\
        \bottomrule
    \end{tabular}
    \label{paraso3}
\end{table}

\newpage
\subsection{Distance metric results in density interpolation}\label{subappendix:dis}
Below the tables are the generation results measured by $\mathrm{K}-\mathrm{L}$ divergence in the experiment of interpolation between mixture distribution on $\mathbb{S}^2$ and $\mathbf{SO}(3)$. The number at row $m$ column $n$ denotes the '$\mathrm{K}-\mathrm{L}$ divergence between the generated $\widehat{\mathbb{V}}^n$ samples from $\mathbb{V}^m$ and the ground truth $\mathbb{V}^n$'(as well as $\mathbb{W}$), where the two numbers in row 32 column 32 are forward/backward results respectively.

\begin{table}[htbp]
    \caption{Generation results from E-SDE scheme \ref{eSDE} in the experiment of interpolation between VMF mixture distribution on $\mathbb{S}^2$. }    
    \label{tab:cwoS2}
    \centering
    \vspace{15pt}
    \begin{tabular}{ccccc}
        \toprule
        centers  & 8 & 16 & 32 & 64\\
        \midrule
        8 & 0 & - & 0.0174& 0.0117 \\
        16 & -& 0 & -  & 0.0140  \\
        32 & 0.0065 & - & 0.0071/0.0161 & - \\
        64 & 0.0335 & 0.0161 & - & 0 \\
        \bottomrule
    \end{tabular}
\end{table}

\begin{table}[htbp]
    \caption{Generation results from ODE scheme \ref{ODE} in the experiment of interpolation between VMF mixture distribution on $\mathbb{S}^2$.}
    \label{tab:cgoS2}
    \centering
    \vspace{15pt}
    \begin{tabular}{ccccc}
        \toprule
        Centers  & 8 & 16 & 32 & 64\\
        \midrule
        8 & 0 & - & 0.0013 & 0.0088 \\
        16 & -& 0 & -  & 0.0198 \\
        32 & 0.0130 & - & 0.0143/0.0254 & - \\
        64 & 0.0067 & 0.0302 & - & 0 \\
        \bottomrule
    \end{tabular}
\end{table}

\begin{table}[htbp]
    \caption{Generation results from GRW scheme \ref{GRW} in the experiment of interpolation between VMF mixture distribution on $\mathbb{S}^2$. }
    \label{tab:cdoS2}
    \centering
    \vspace{15pt}
    \begin{tabular}{ccccc}
        \toprule
        Centers  & 8 & 16 & 32 & 64\\
        \midrule
        8 & 0 & - & 0.4586 & 0.2885 \\
        16 & -& 0 & -  & 0.1976 \\
        32 & 0.3436 & - & 0.1155/0.0892 & - \\
        64 & 0.5521 & 0.1907 & - & 0 \\
        \bottomrule
    \end{tabular}
\end{table}

\begin{table}[htbp]
    \vspace{-10pt}
    \caption{Generation results from ODE scheme \ref{ODE} of RNGI-D model in the experiment of interpolation between wrapped Gaussian mixture distribution on $\mathbf{SO}(3)$. }
    \label{tab:cdoDo}
    \centering
    \vspace{10pt}
    \begin{tabular}{ccccc}
        \toprule
        Centers  & 8 & 16 & 32 & 64\\
        \midrule
        8 & 0 & - & 0.00091 &  0.00083 \\
        16 & -& 0 & -  &  0.00098 \\
        32 & 0.00005 & - & 0.00241/0.00024 & - \\
        64 & 0.00003 & 0.00017 & - & 0 \\
        \bottomrule
    \end{tabular}
\end{table}

\begin{table}[htbp]
    \vspace{-10pt}
    \caption{Generation results from GRW scheme \ref{GRW} of RNGI-D model in the experiment of interpolation between wrapped Gaussian mixture distribution on $\mathbf{SO}(3)$.}
    \label{tab:cdoDg}
    \centering
    \vspace{10pt}
    \begin{tabular}{ccccc}
        \toprule
        Centers  & 8 & 16 & 32 & 64\\
        \midrule
        8 & 0 & - & 0.00463 & 0.00360 \\
        16 & -& 0 & -  & 0.00116 \\
        32 & 0.00088 & - & 0.00278/0.00092 & -\\
        64 & 0.00059 & 0.00168 & - & 0 \\
        \bottomrule
    \end{tabular}
\end{table}

\begin{table}[htbp]
    \vspace{-10pt}
    \caption{Generation results from ODE scheme \ref{ODE} of RNGI-E model in the experiment of interpolation between wrapped Gaussian mixture distribution on $\mathbf{SO}(3)$.}
    \label{tab:cdoEo}
    \centering
    \vspace{10pt}
    \begin{tabular}{ccccc}
        \toprule
        Centers  & 8 & 16 & 32 & 64\\
        \midrule
        8 & 0 & - & 0.00336 & 0.00036 \\
        16 & -& 0 & -  & 0.00093 \\
        32 & 0.00021 & - & 0.00320/0.00071 & - \\
        64 & 0.00013 & 0.00043 & - & 0 \\
        \bottomrule
    \end{tabular}
\end{table}

\begin{table}[htbp]
    \vspace{-10pt}
    \caption{Generation results from E-SDE scheme \ref{eSDE} of RNGI-E model in the experiment of interpolation between wrapped Gaussian mixture distribution on $\mathbf{SO}(3)$. }
    \label{tab:cdoEs}
    \centering
    \vspace{10pt}
    \begin{tabular}{ccccc}
        \toprule
        Centers  & 8 & 16 & 32 & 64\\
        \midrule
        8 & 0 & - & 0.01802 & 0.00513 \\
        16 & -& 0 & -  & 0.00408 \\
        32 & 0.00102 & - & 0.01073/0.00139 & - \\
        64 & 0.00088 & 0.00684 & - & 0 \\
        \bottomrule
    \end{tabular}
\end{table}

\newpage
\subsection{Visualization}\label{subappendix:vis}
\begin{figure}[htbp]
    \centering
    \begin{subfigure}[b]{0.48\textwidth}
      \includegraphics[width=\textwidth]{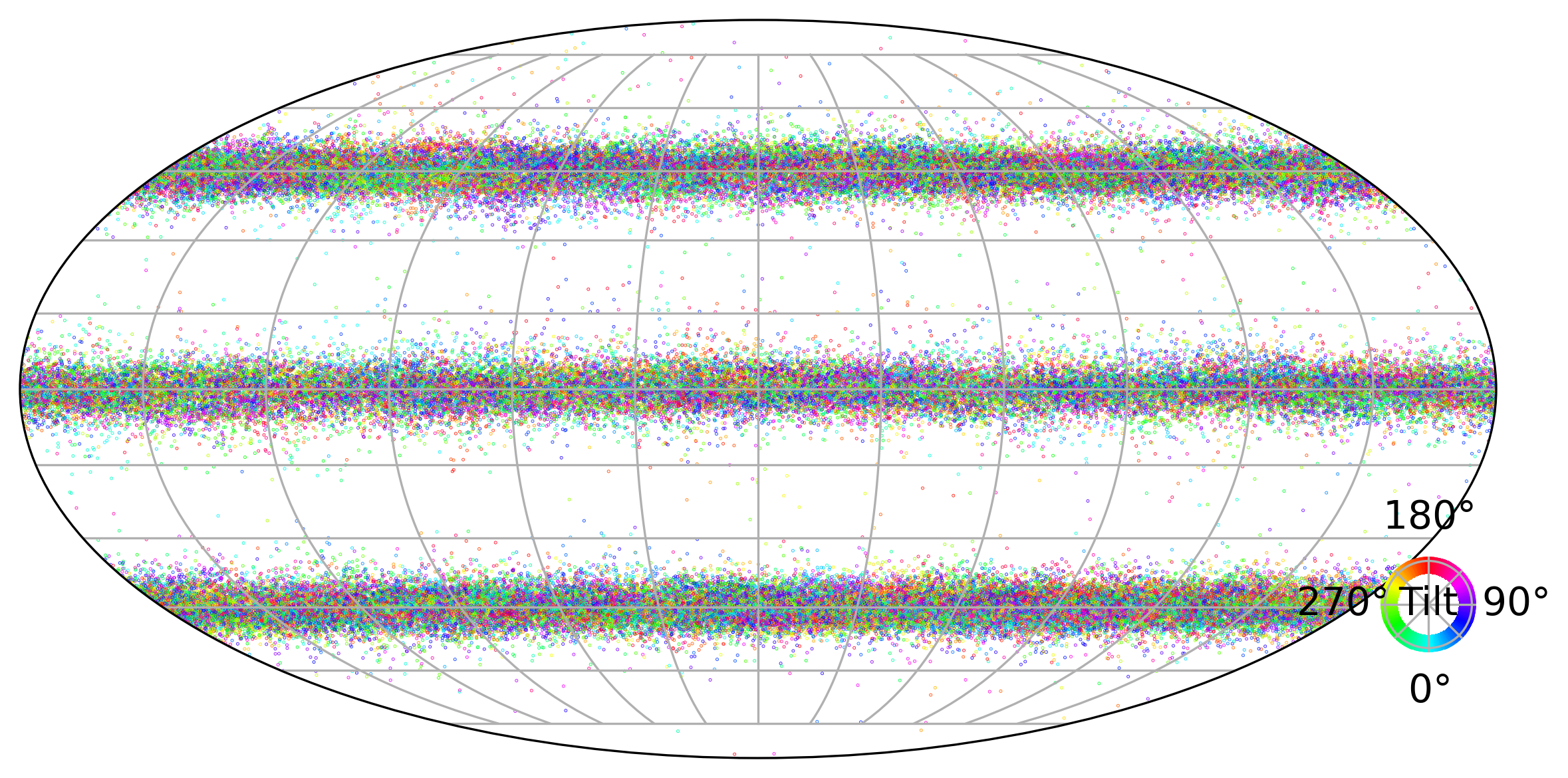}
      \caption{}
      \label{fig:lineEMo}
    \end{subfigure}
    ~% add desired spacing
    \begin{subfigure}[b]{0.48\textwidth}
      \includegraphics[width=\textwidth]{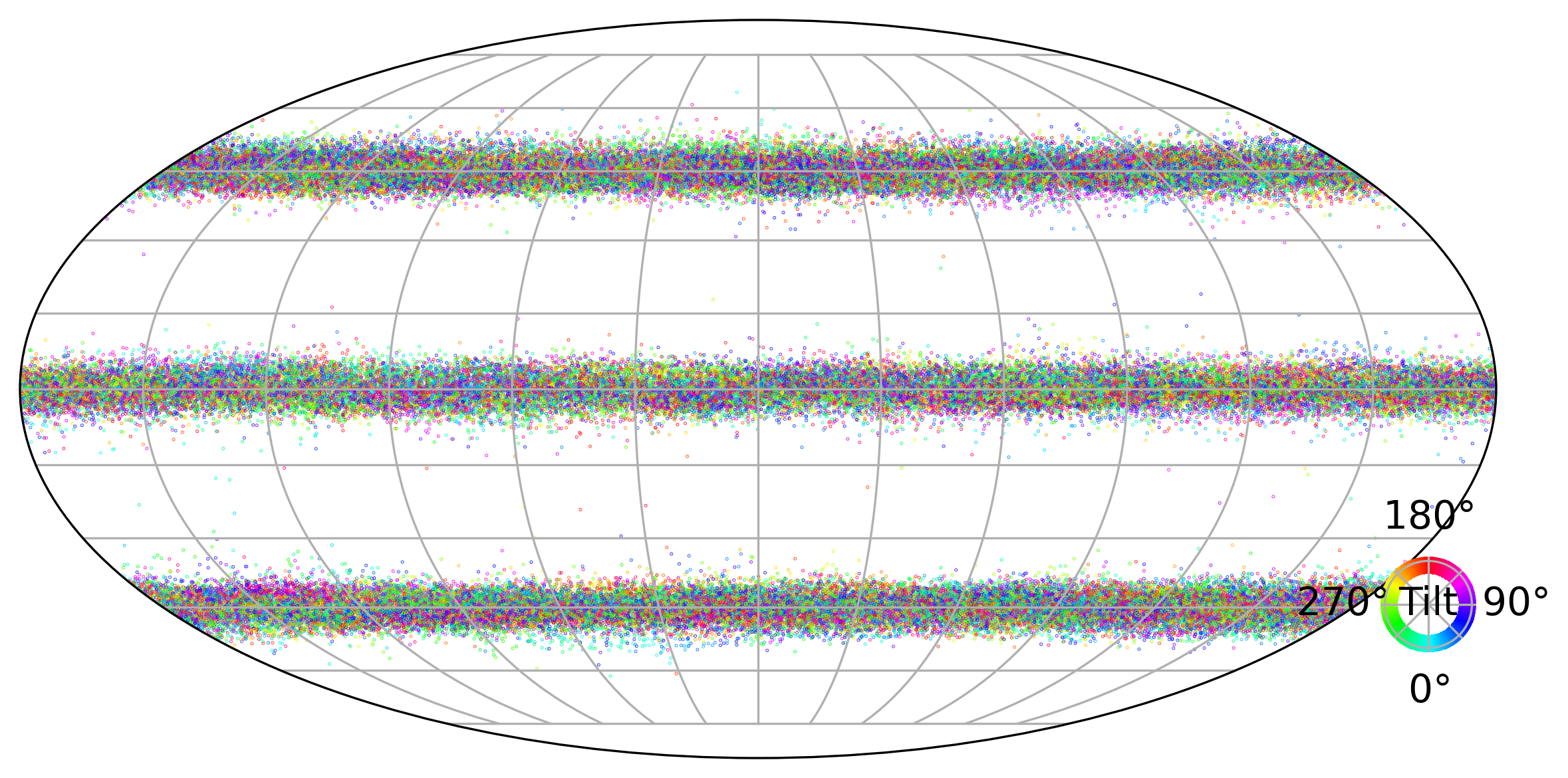}
      \caption{}
      \label{fig:lineESo}
    \end{subfigure}
    \\% line break
    \begin{subfigure}[b]{0.48\textwidth}
      \includegraphics[width=\textwidth]{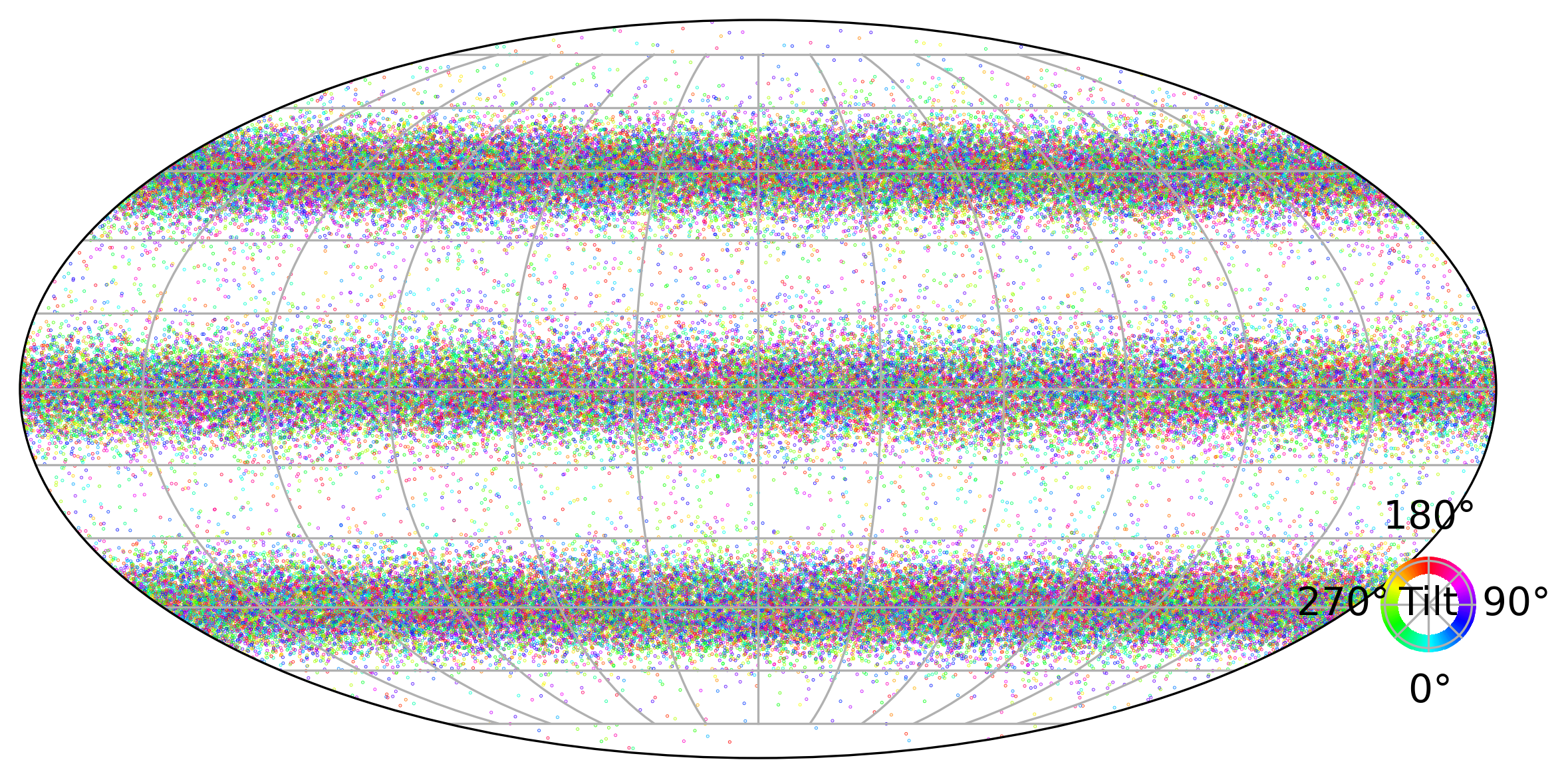}
      \caption{}
      \label{fig:lineEMs}
    \end{subfigure}
    ~% add desired spacing
    \begin{subfigure}[b]{0.48\textwidth}
      \includegraphics[width=\textwidth]{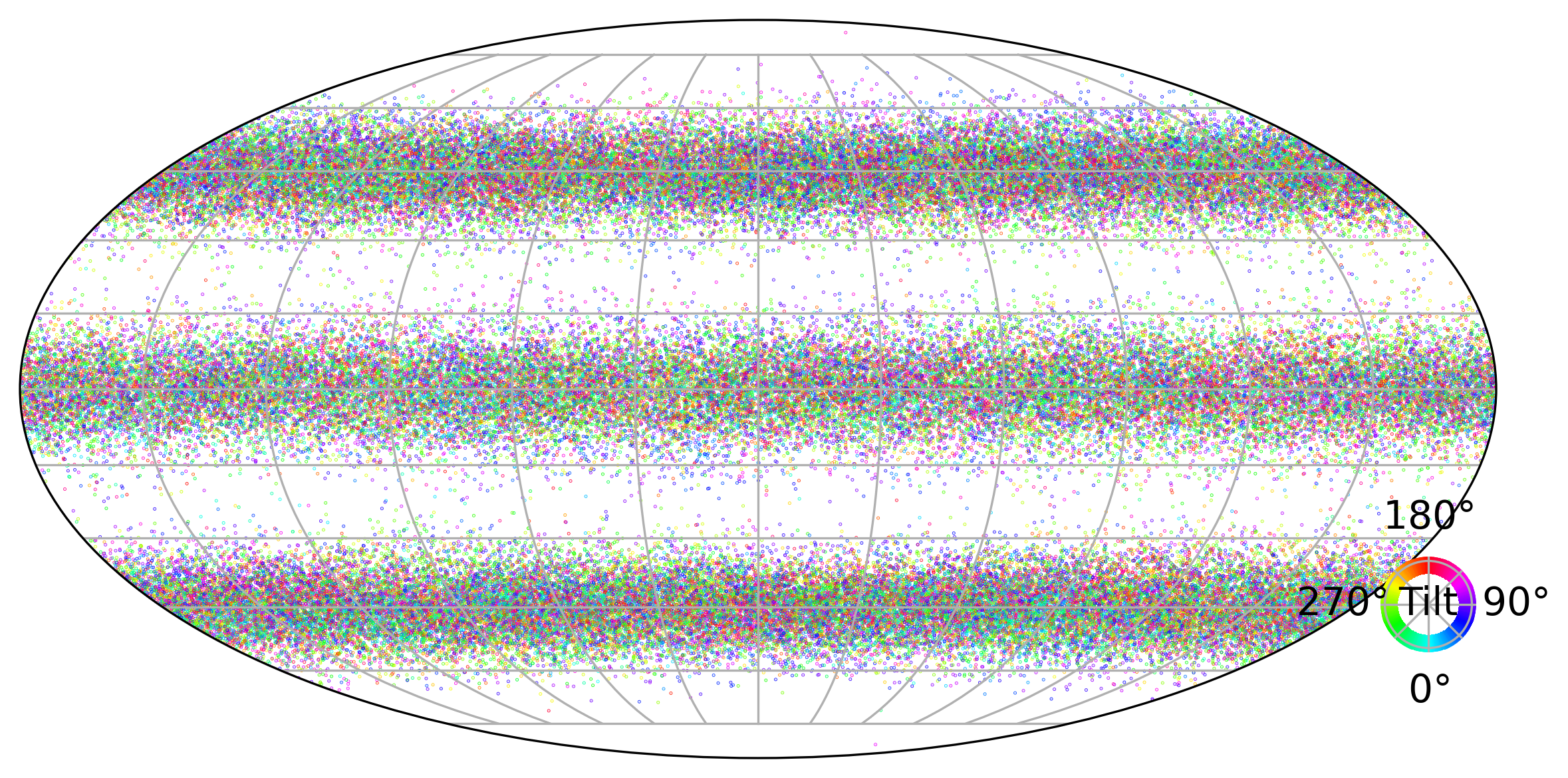}
      \caption{}
      \label{fig:lineESs}
    \end{subfigure}
    \caption{\enspace Generative performance of RNGI models on the line dataset of density generation experiment(1) on $\mathbf{SO}(3)$.}
    \label{fig:line4}
\end{figure}

\begin{figure}[b]
    \centering
    \begin{subfigure}[b]{0.48\textwidth}
      \includegraphics[width=\textwidth]{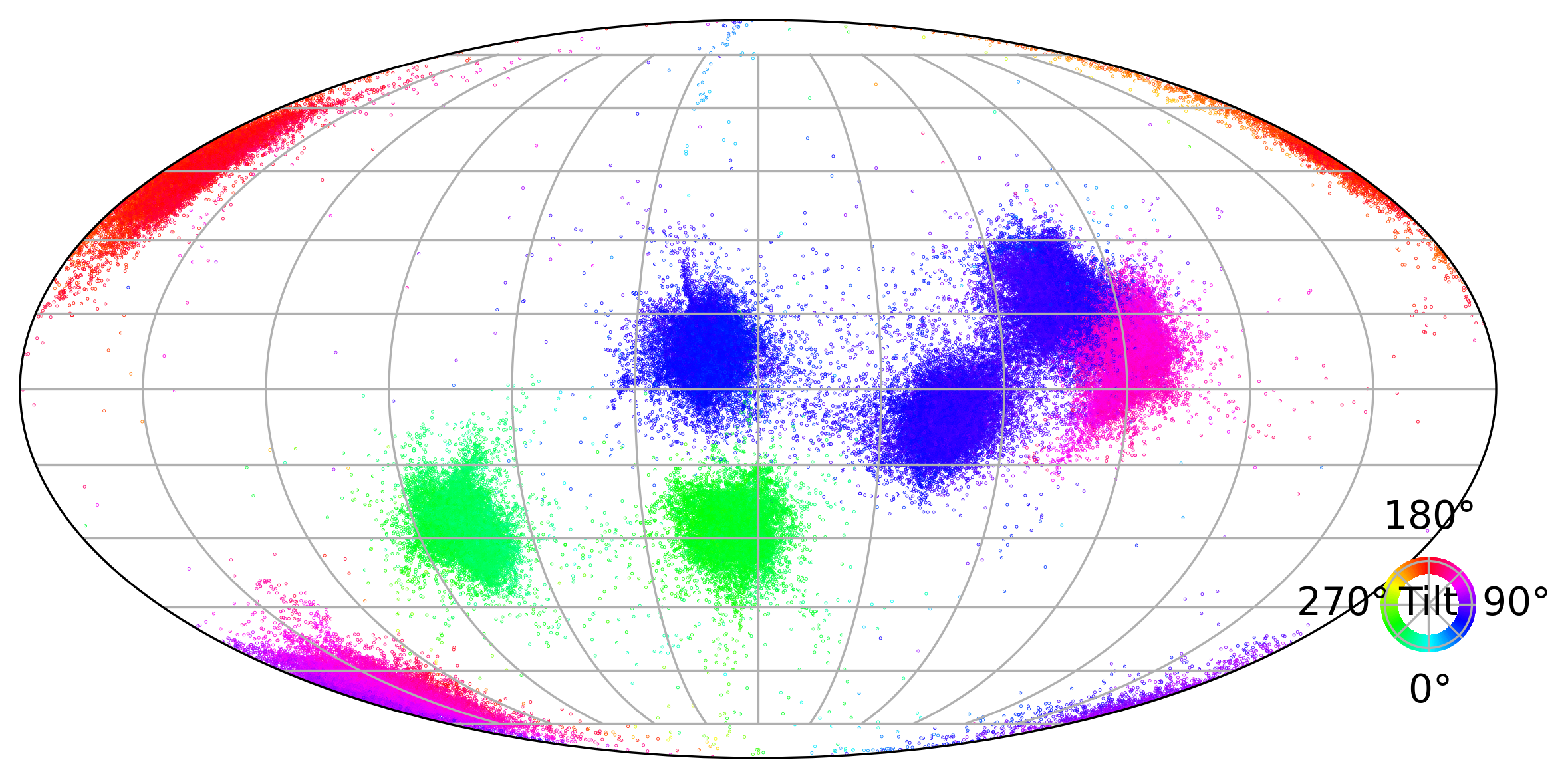}
      \caption{}
      \label{fig:328o}
    \end{subfigure}
    ~% add desired spacing
    \begin{subfigure}[b]{0.48\textwidth}
      \includegraphics[width=\textwidth]{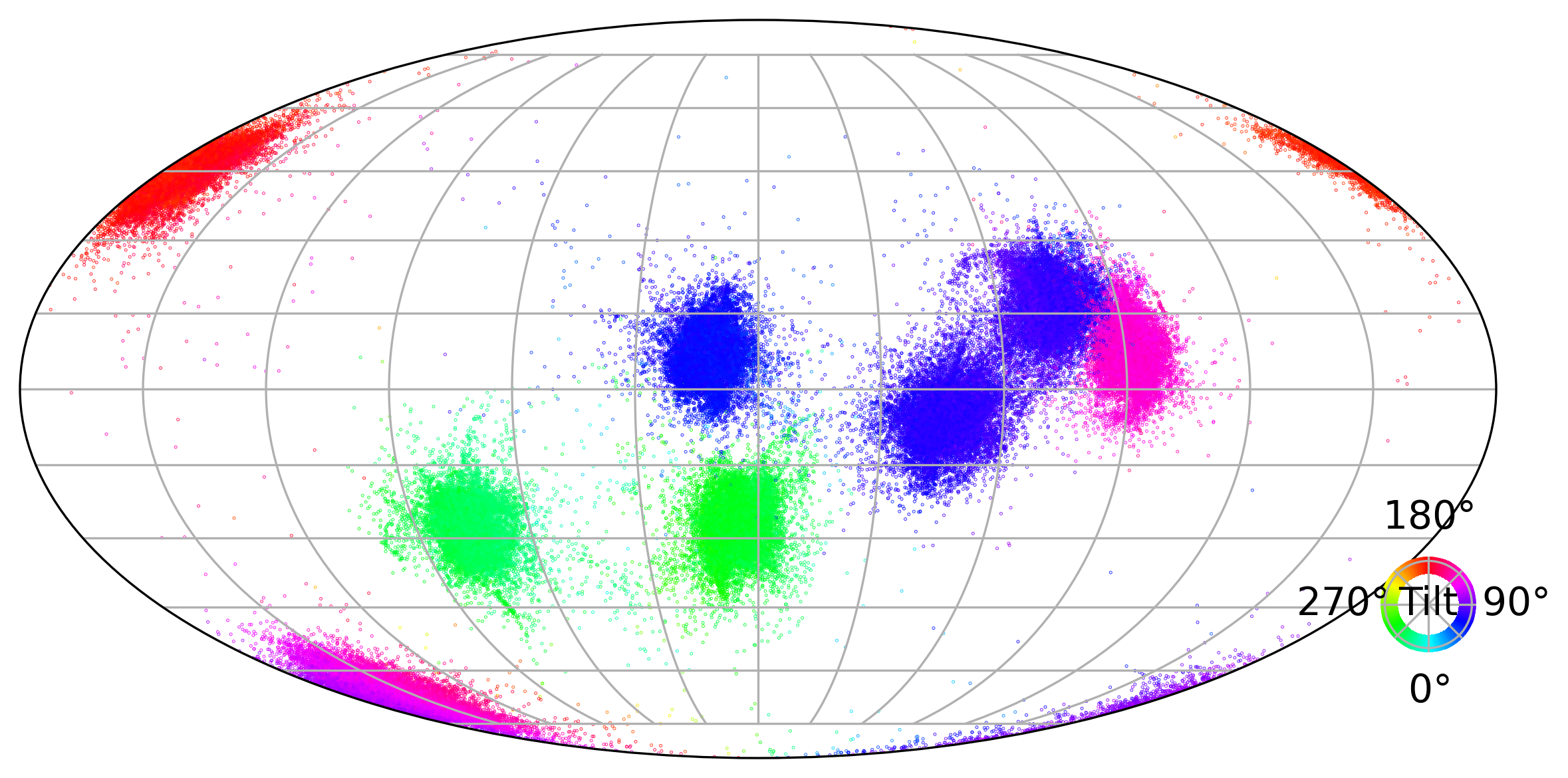}
      \caption{}
      \label{fig:648o}
    \end{subfigure}
    \\% line break
    \begin{subfigure}[b]{0.48\textwidth}
      \includegraphics[width=\textwidth]{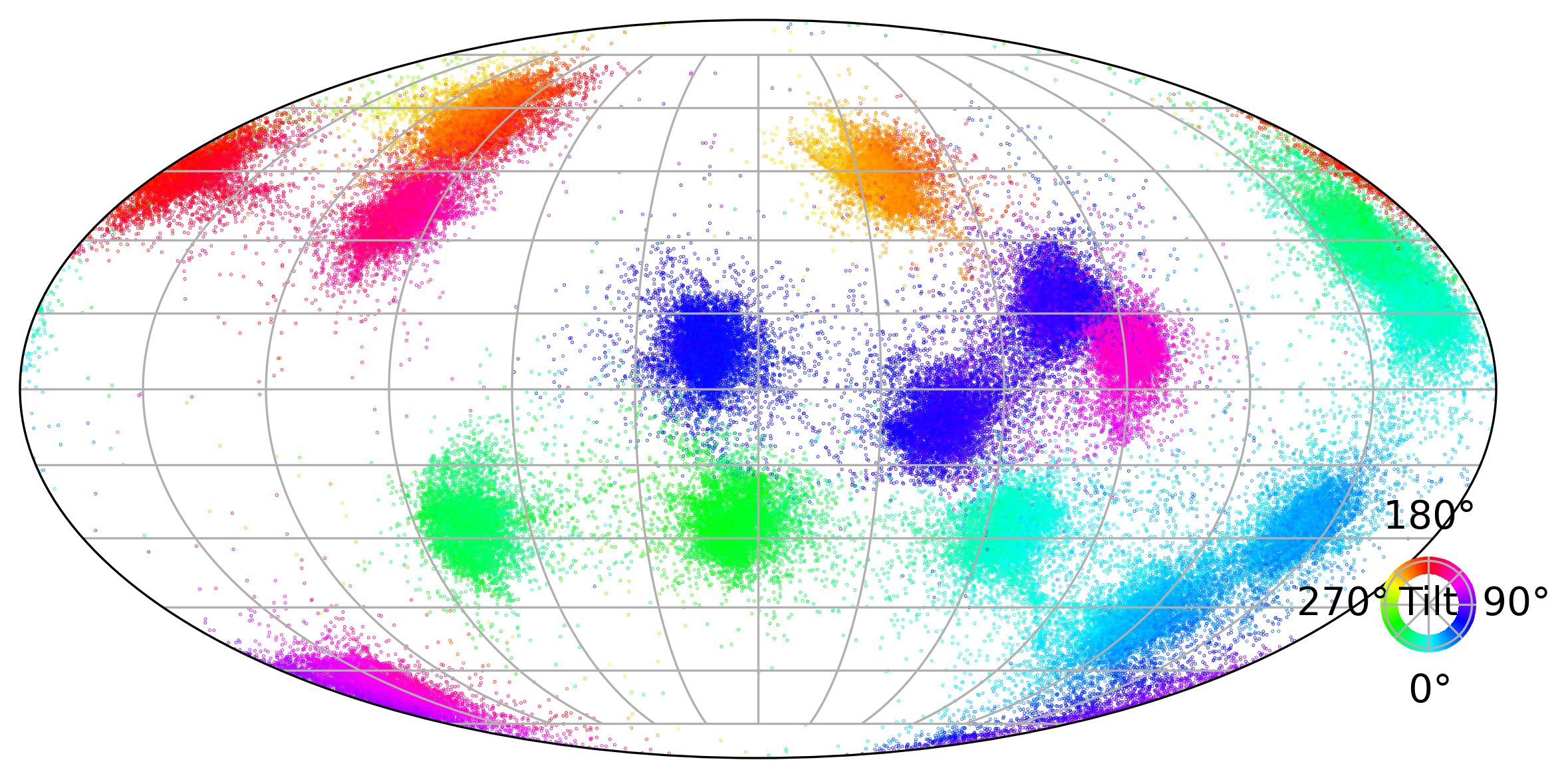}
      \caption{}
      \label{fig:6416o}
    \end{subfigure}
    ~% add desired spacing
    \begin{subfigure}[b]{0.48\textwidth}
      \includegraphics[width=\textwidth]{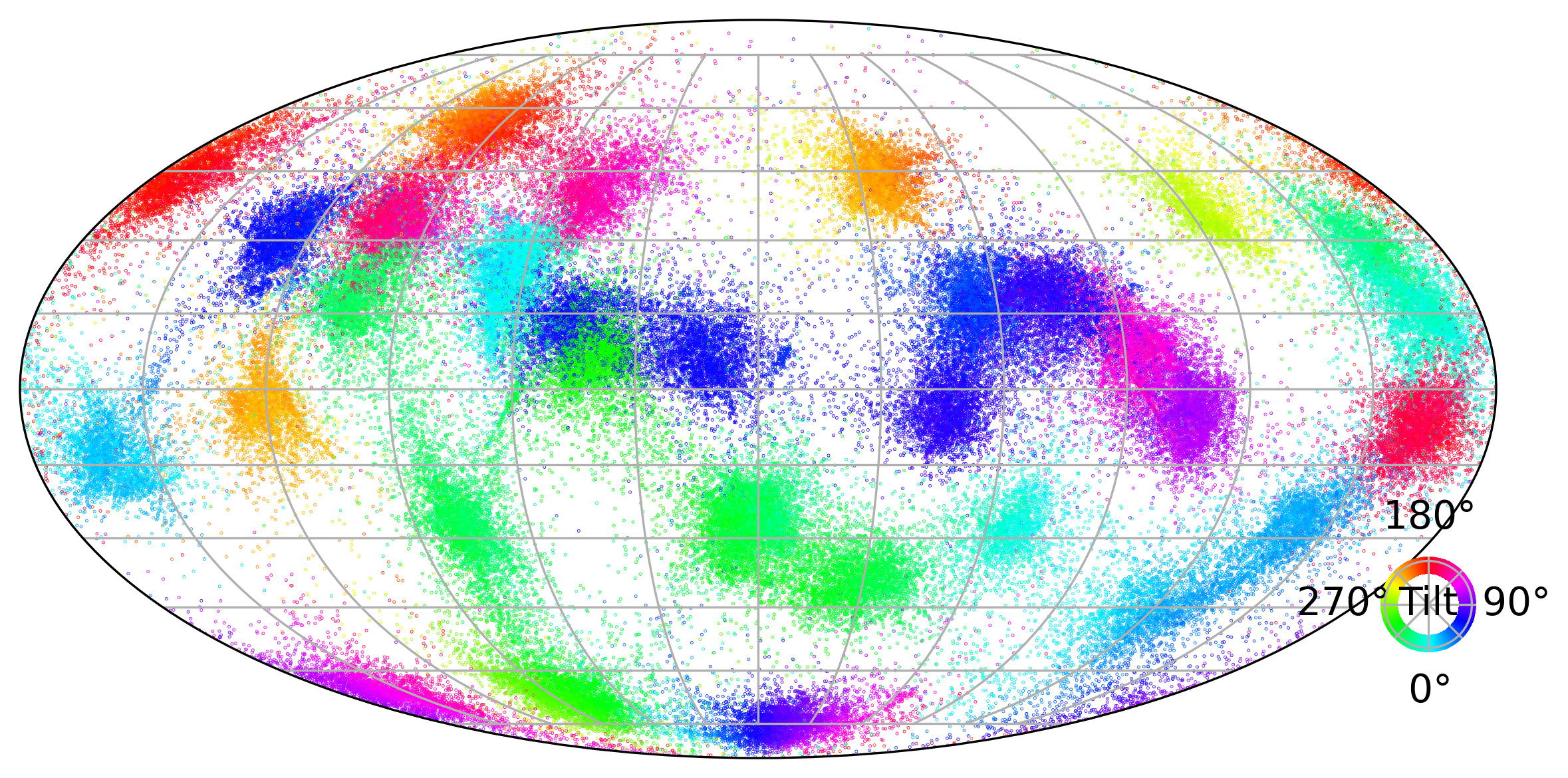}
      \caption{}
      \label{fig:3232o}
    \end{subfigure}
    \caption{Generative performance of RNGI-E models by its ODE scheme \eqref{ODE} in the distribution connection experiment on $\mathbf{SO}(3)$.}
    \label{fig:0o1}
\end{figure}

\begin{figure}[b]
    \centering
    \begin{subfigure}[b]{0.48\textwidth}
      \includegraphics[width=\textwidth]{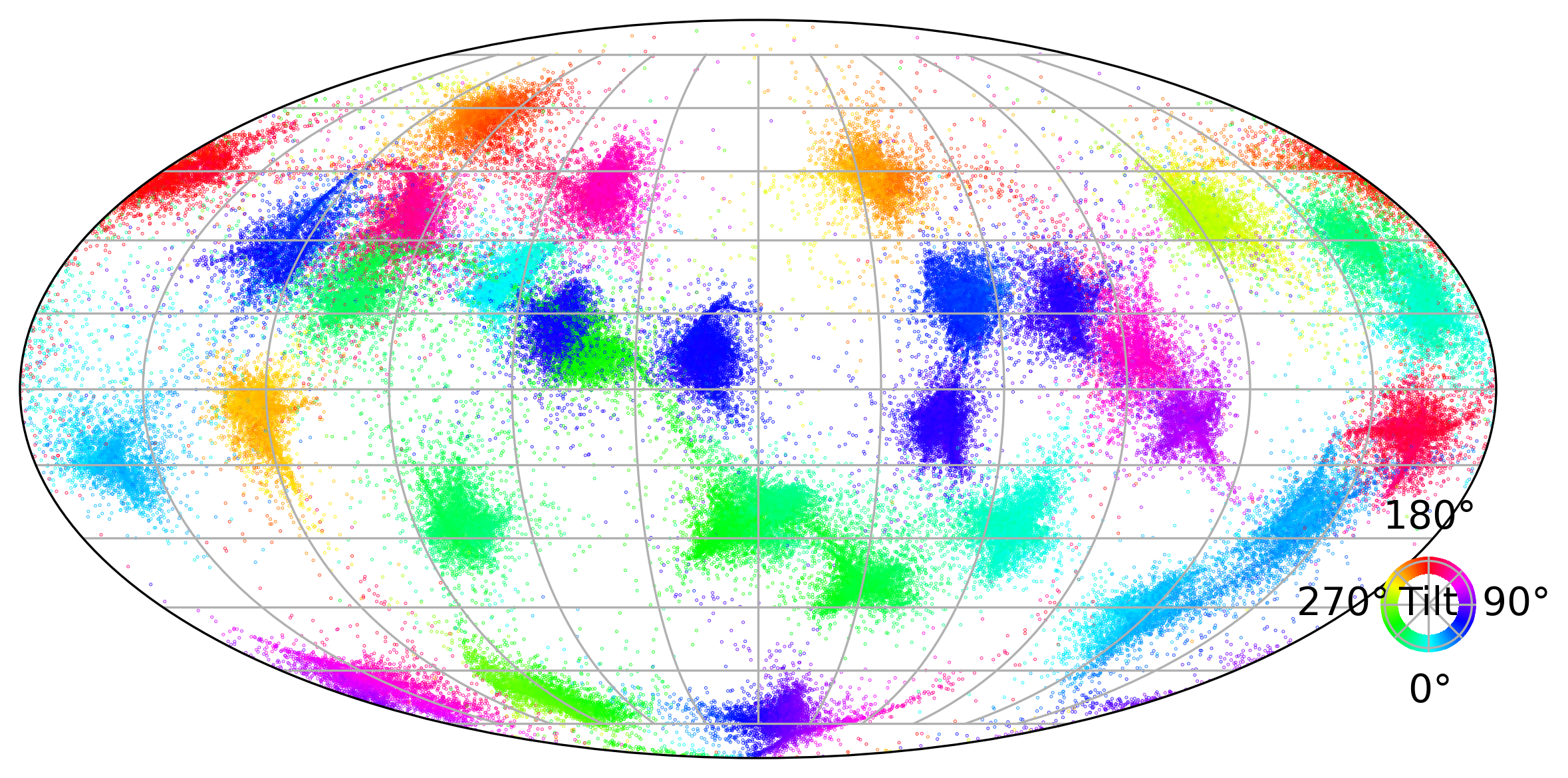}
      \caption{}
      \label{fig:328doi}
    \end{subfigure}
    ~% add desired spacing
    \begin{subfigure}[b]{0.48\textwidth}
      \includegraphics[width=\textwidth]{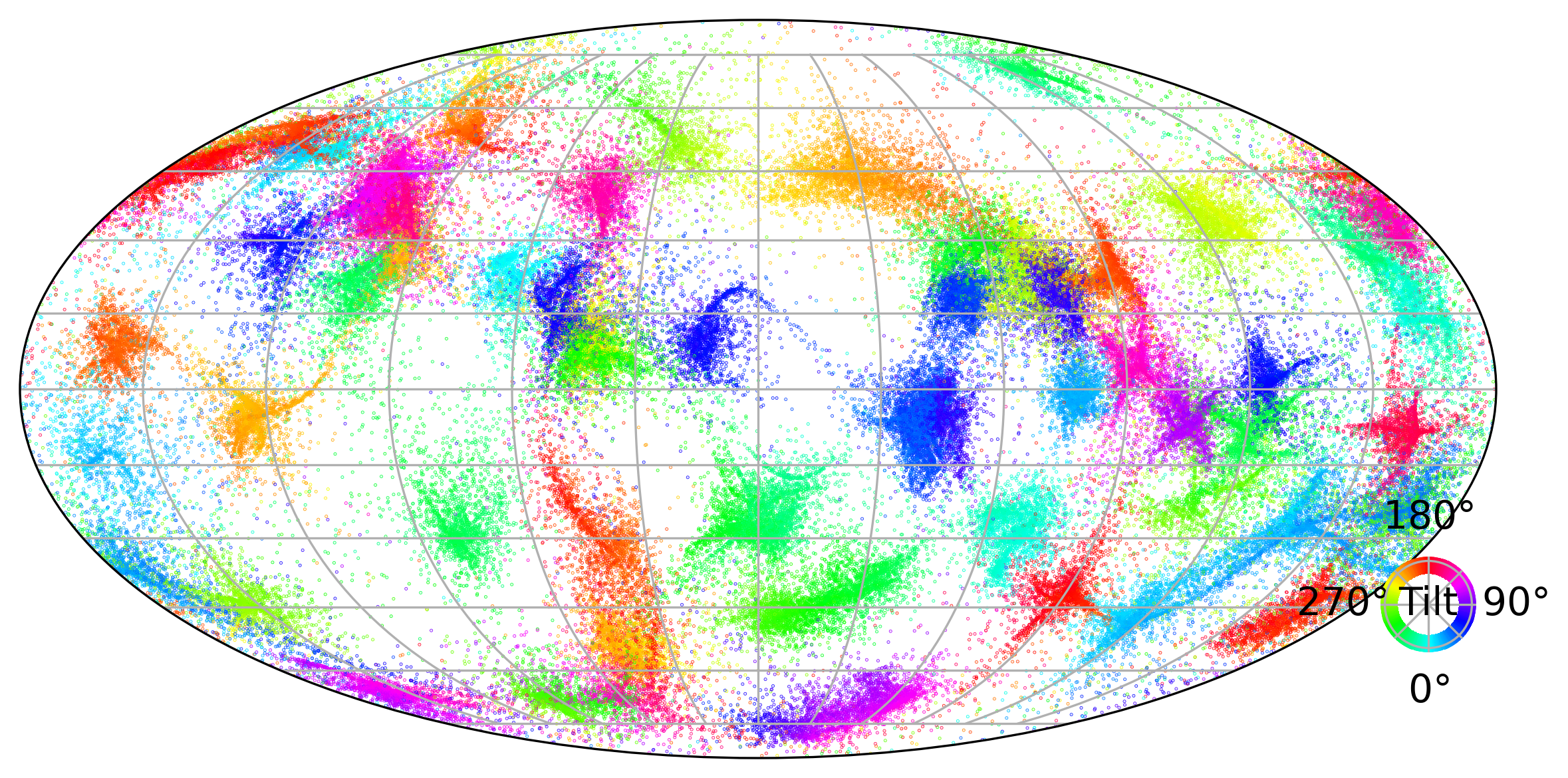}
      \caption{}
      \label{fig:648doi}
    \end{subfigure}
    \\% line break
    \begin{subfigure}[b]{0.48\textwidth}
      \includegraphics[width=\textwidth]{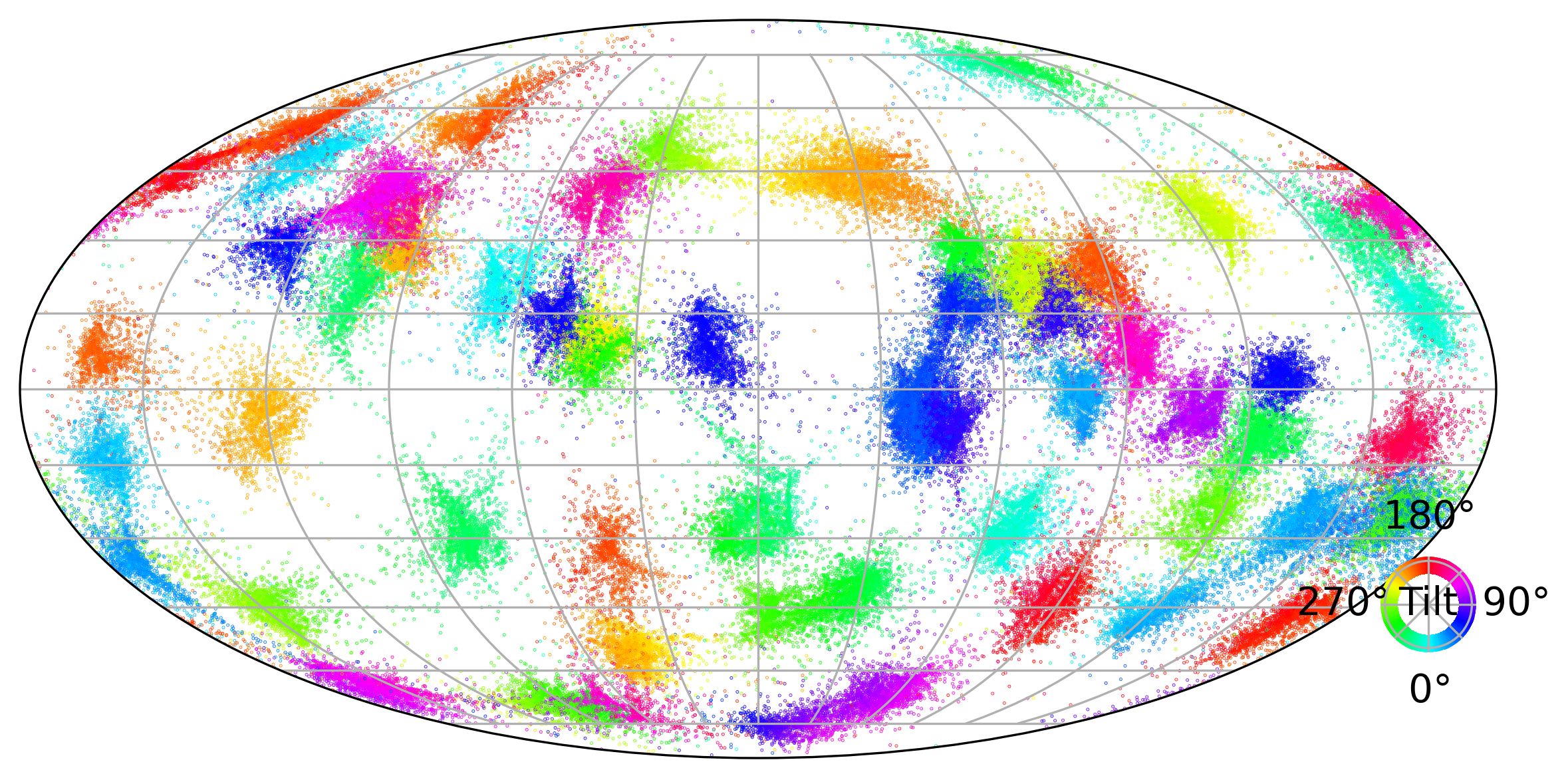}
      \caption{}
      \label{fig:6416doi}
    \end{subfigure}
    ~% add desired spacing
    \begin{subfigure}[b]{0.48\textwidth}
      \includegraphics[width=\textwidth]{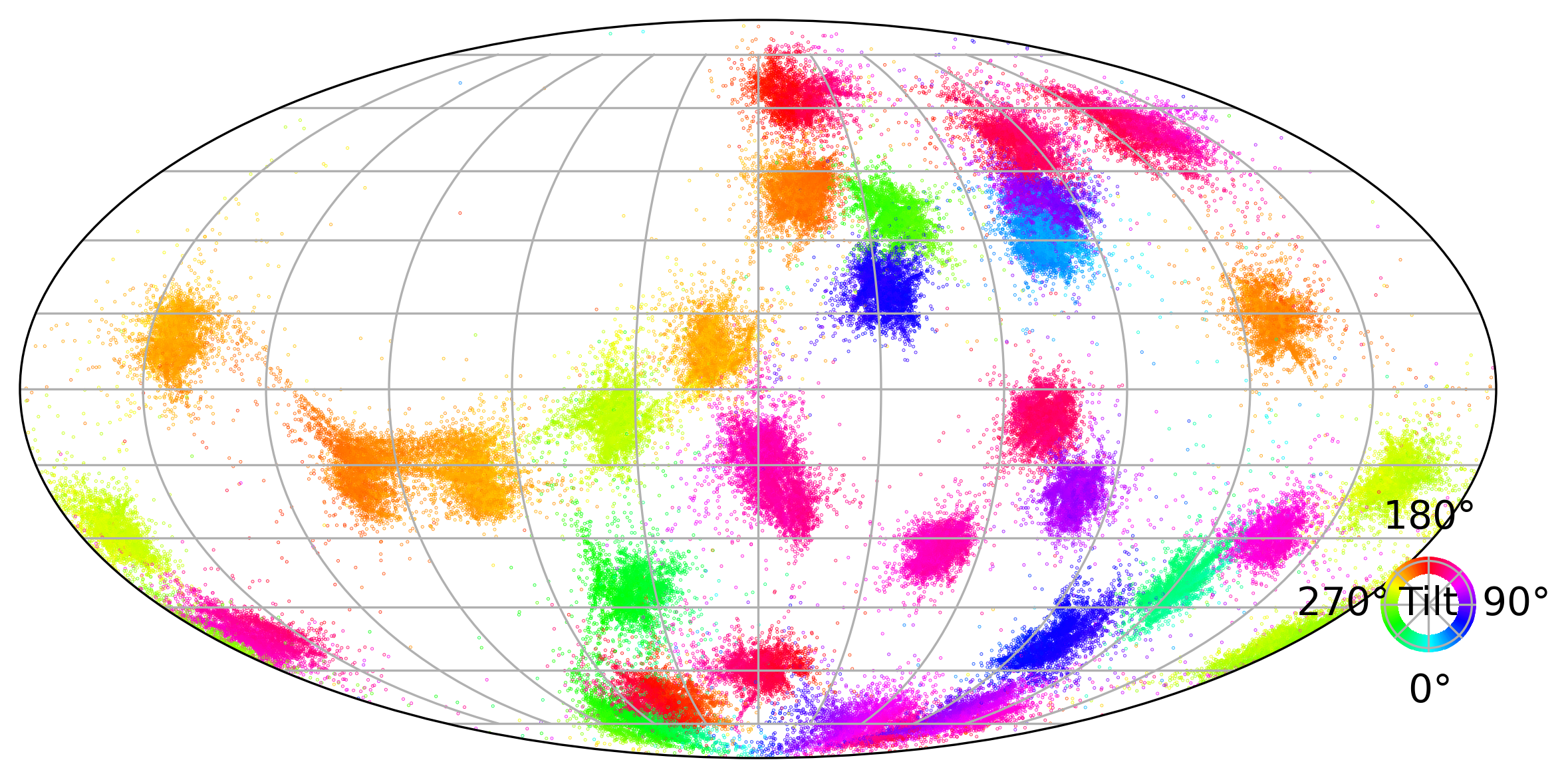}
      \caption{}
      \label{fig:3232doi}
    \end{subfigure}
    \caption{\enspace Generative performance of RNGI-D models by its ODE scheme \eqref{ODE} in the distribution connection experiment on $\mathbf{SO}(3)$.}
    \label{fig:d1o0}
\end{figure}

Below the figures are the generative visualization of RNGI models in both density generation experiments and density interpolation experiments. The order of subfigures in Figure \ref{fig:line4} is EM-ODE, ES-ODE, EM-ESDE and ES-ESDE, and that in Figure \ref{fig:0o1} is $\mathbb{W}^{32}\to\mathbb{W}^8$, $\mathbb{W}^{64}\to\mathbb{W}^8$, $\mathbb{W}^{64}\to\mathbb{W}^{16}$ and $\mathbb{W}^{32}_{\{\mu_i\}}\to\mathbb{W}^{32}_{\{\nu_i\}}$ respectively, while the order in Figure \ref{fig:d1o0} is the same but the generative directions are reversed. 

\end{document}